\documentclass{article}

\usepackage[final, nonatbib]{neurips_2020}

\usepackage[utf8]{inputenc} % allow utf-8 input
\usepackage[T1]{fontenc}    % use 8-bit T1 fonts
\usepackage{hyperref}       % hyperlinks
\usepackage{url}            % simple URL typesetting
\usepackage{booktabs}       % professional-quality tables
\usepackage{amsfonts}       % blackboard math symbols
\usepackage{nicefrac}       % compact symbols for 1/2, etc.
\usepackage{microtype}      % microtypography

\usepackage{caption}
\usepackage{graphicx}
\usepackage{color}
\usepackage{amsmath}
\usepackage{titlesec}
\usepackage{subcaption}

\titleformat{\paragraph}
{\normalfont\normalsize\bfseries}{\theparagraph}{1em}{}
\titlespacing*{\paragraph}{0pt}{3.25ex plus 1ex minus .2ex}{1.5ex plus .2ex}

\title{Unsupervised Domain Adaptation Approaches for Chessboard Recognition}

\author{
    Wassim Jabbour \\
  McGill University\\
  \texttt{wassim.jabbour@mail.mcgill.ca}
  \And
  Enzo Benoit-Jeannin \\
  McGill University \\
  \texttt{enzo.benoit-jeannin@mail.mcgill.ca}
  \AND
  Oscar Bedford \\
  McGill University \\
  \texttt{oscar.bedford@mail.mcgill.ca}
  \And
  Saif Shahin \\
  McGill University \\
  \texttt{saif.shahin@mail.mcgill.ca}
}

\begin{document}

\maketitle
\begin{abstract}
Chess involves extensive study and requires players to keep manual records of their matches, a process which is time-consuming and distracting. The lack of high-quality labeled photographs of chess boards, and the tediousness of manual labeling, have hindered the wide application of Deep Learning (DL) to automating this record-keeping process. This paper proposes an end-to-end pipeline that employs domain adaptation (DA) to predict the labels of real, top-view, unlabeled chessboard images using synthetic, labeled images. The pipeline is composed of a pre-processing phase which detects the board, crops the individual squares, and feeds them one at a time to a DL model. The model then predicts the labels of the squares and passes the ordered predictions to a post-processing pipeline which generates the Forsyth–Edwards Notation (FEN) of the position. The three approaches considered are the following: A VGG16 model pre-trained on ImageNet, defined here as the Base-Source model, fine-tuned to predict source domain squares and then used to predict target domain squares without any domain adaptation; an improved version of the Base-Source model which applied CORAL loss to some of the final fully connected layers of the VGG16 to implement DA; and a Domain Adversarial Neural Network (DANN) which used the adversarial training of a domain discriminator to perform the DA. Also, although we opted not to use the labels of the target domain for this study, we trained a baseline with the same architecture as the Base-Source model (Named Base-Target) directly on the target domain in order to get an upper bound on the performance achievable through domain adaptation. The results show that the DANN model only results in a 3\% loss in accuracy when compared to the Base-Target model while saving all the effort required to label the data.
\end{abstract}

\section{Introduction}
Chess, a centuries-old strategic board game, remains popular among enthusiasts and masters alike. In long-format competitive play, every move executed by each side must be manually recorded using pen and paper. One reason for this is post-game analysis, where chess players will review their games to improve their strategies and skills. Another reason is to resolve conflict about illegal moves or to determine draws by number of repetitions, among others. The need for an automatic record of chess moves on physical boards emerges from the time-consuming nature of manual record keeping for both long-format and short-format games. While this is an area where Deep Learning can be applied, there has not been much progress. This can be attributed to the difficulty of gathering enough labeled data, as there are up to 32 pieces and associated locations to label per image. 
This paper proposes an unsupervised approach to automatic annotation of chessboard photographs. The approach centers around the concept of unsupervised domain adaptation, a technique used to improve the performance of a model on a target domain containing no labeled data by using the knowledge learned by the model from a related source domain with an abundance of labeled data with a slightly different distribution {\cite{DA_OG}}. The source domain data employed to perform the domain adaptation consists of 3D images of chessboards rendered using Blender \cite{blender}, as these are simple to generate in large numbers and can be designed to match the distribution of the target domain. The target domain data employed was unlabeled top-view photographs of chess positions.

From a broader perspective, the proposed solution consists of 3 components: A pre-processing pipeline which takes a full target domain photograph as input, detects the board, and crops out the individual squares. Then, the individual squares are passed one at a time to a Deep Learning model trained using domain adaptation which can classify the labels of the chess pieces on each square. Finally, the ordered predictions of the model are passed to a post-processing pipeline which generates a FEN string representing the position that can be fed to a chess engine to generate a complete 2D representation of the input.

In order for the proposed Domain Adaptation model to be successful, there are several crucial constraints that must be addressed. Firstly, the 3D generated data must closely resemble target domain photographs in terms of camera angle, lighting, board colors, and piece textures, ensuring that the extent of domain adaptation is minimized. Secondly, this project is circumscribed to automating the annotation of individual images, rather than full videos. Finally, on a higher level, a goal-oriented constraint is that the model is expected to perform significantly better in classifying the target domain squares than a model trained on the source data and directly tested on the target data without any domain adaptation.

This project's synthetic 3D chess images and the domain adaptation model are freely available for non-commercial use. Source code is provided under an open-source license at \href{https://github.com/WassimJabz/RecogniChess}{https://github.com/WassimJabz/RecogniChess}, implemented in Python, and supported on major platforms. Additional figures, data, and methodology can be accessed in the same repository.

\section{Methodology}

\subsection{Datasets}

\subsubsection{Target dataset}

\begin{figure}[!ht]
\centering
\includegraphics[scale=0.5]{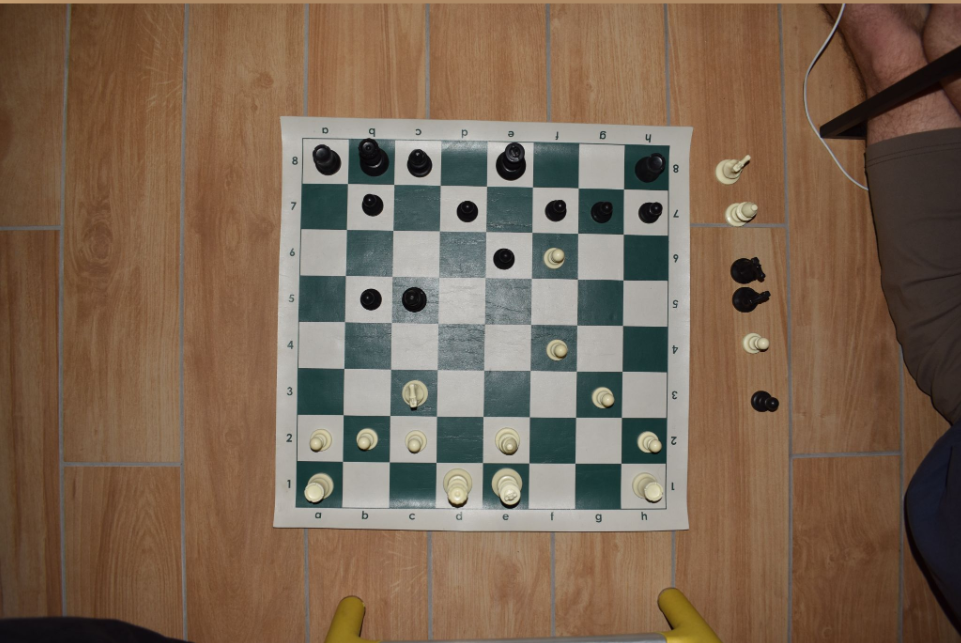}
\caption{Example of an image taken from the target dataset.}
\label{fig:reallifeexample}
\end{figure}

The target dataset comprises 500 photographs of actual chess positions displayed on a chessboard, as demonstrated in Figure \ref{fig:reallifeexample}. These images were made available thanks to an open-source effort by fellow chess enthusiasts {\cite{ChessDataset}}. The images purposefully include a noisy background around the board to simulate real-life situations where getting clear images of the board might not always be possible, especially in short format games. A pre-processing pipeline using Computer Vision was designed to detect the board in the image, warp it to top-view, and crop out individual squares. This process will be explained in Section {\ref{subsubsection:Target_preprocessing}}.

It is to be noted that, while labels of the positions were provided with this dataset, they were discarded during model training in order to simulate an unsupervised domain adaptation situation. However, the  labels were utilized during validation and testing in order to successfully tune the hyperparameters and evaluate the performance of the model, as will be expanded on in Section {\ref{Models}}. Furthermore, the Base-Target baseline model was directly trained on the target domain using those labels to get an upper bound on the performance achievable through DA.

We would like to thank the creators, Kurt Convey, Michael Deng, Samuel Ryan and Mukund Venkateswaran, for their contribution to chess position recognition development. 

\subsubsection{Source dataset generation}
\begin{figure}[!ht]
    \centering
    \makebox[\textwidth][c]{%
    \begin{minipage}{2\textwidth}
        \centering
        \subfloat[Modification of the knight's crest.]{\includegraphics[scale=0.32]{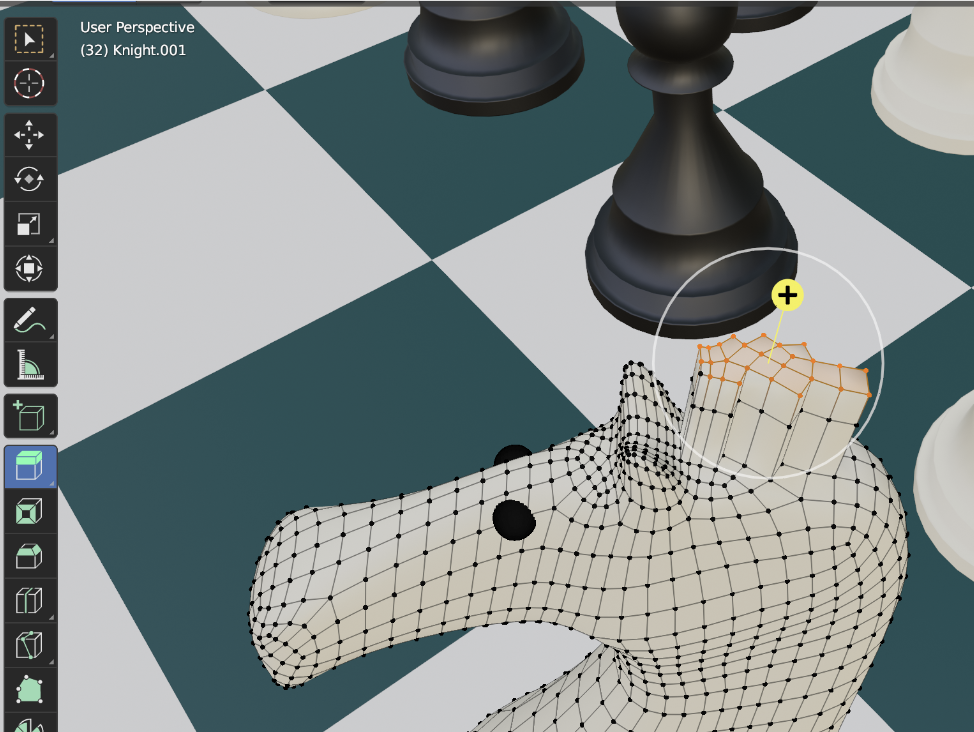}}
        \qquad
        \subfloat[Final Bishop 3D model.]{\includegraphics[scale=0.29]{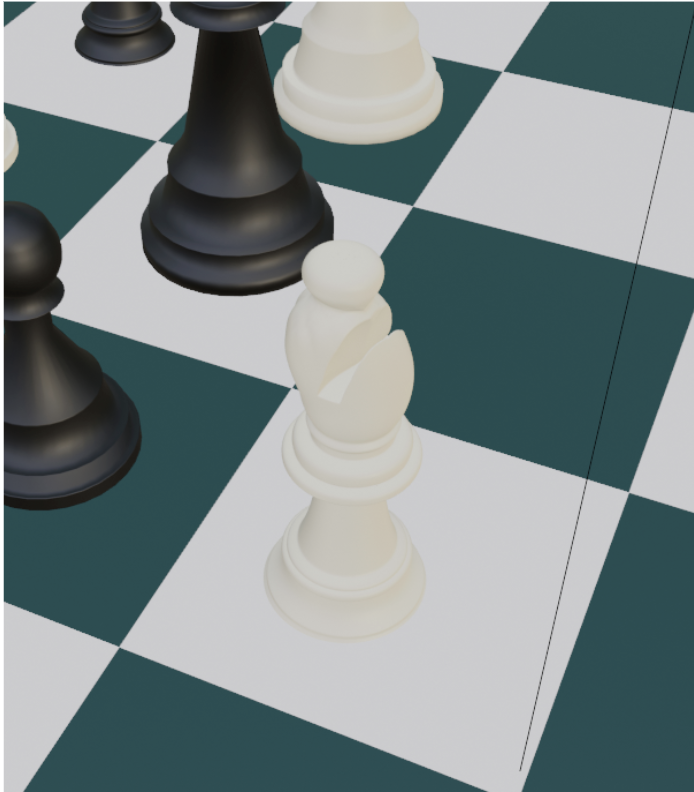}}
        \qquad
        \subfloat[Final Rook 3D model.]{\includegraphics[scale=0.32]{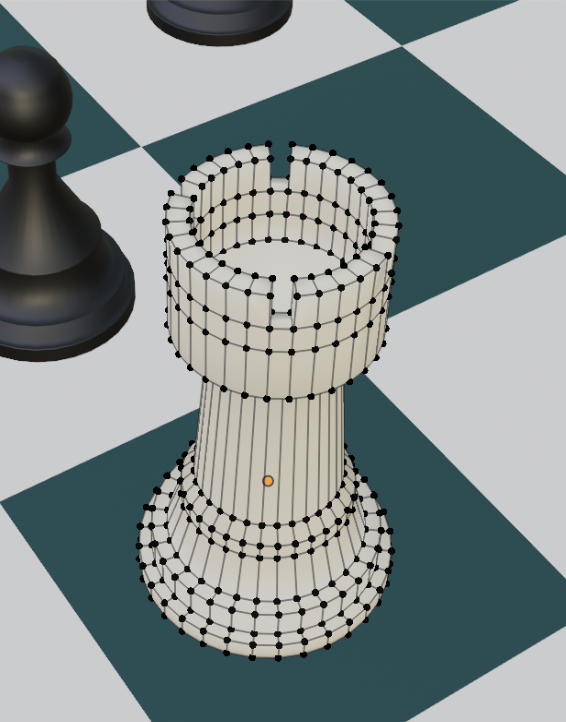}}
    \end{minipage}}
    \caption{3D Chess Piece Modifications in Blender.}
    \label{fig:PieceModifications}
\end{figure}

\begin{figure}[!ht]
\centering
\includegraphics[scale=0.7]{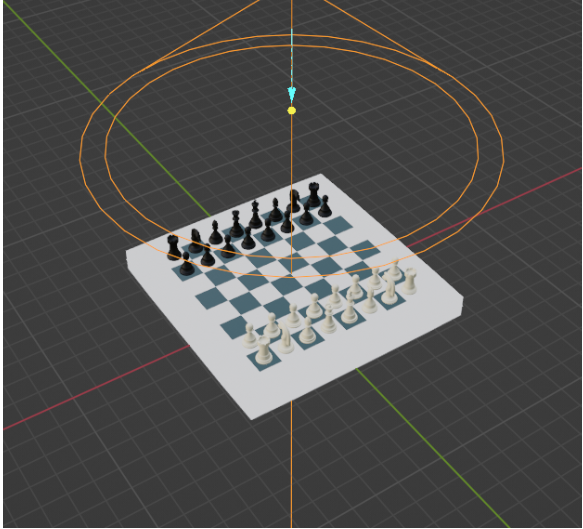}
\caption{Final Blender model of the pieces.}
\label{fig:Chess3D}
\end{figure}

\paragraph{Blender modeling and Python Script}
The source dataset is comprised of 288,000 distinct images of chess squares extracted from a singular virtual chessboard. These images were generated using the open-source 3D-rendering software, Blender \cite{blender}. The pre-designed set of 3D chess pieces, including the board, were obtained from an online source \cite{3dDataset} and imported into the software. Using Blender, the texture of both the chessboard and the chess pieces was customized to replicate the real-life dataset. This involved transforming the white pieces to a subtle beige hue, the black pieces to a darker shade, and the chessboard to a combination of green and white tones. Certain chess pieces from the imported pre-designed set exhibited noticeable visual differences when compared to their real-life counterparts from a top-view perspective. In order to rectify this aesthetic, more suitable pieces were imported from other online sources \cite{knight,bishop}, which were further refined using the 3D tools available in Blender. This is shown for a knight piece in Figure \ref{fig:PieceModifications}a. Similarly, a separate bishop piece was imported from another online source \cite{bishop}, and used after re-scaling and re-shaping its top form as shown on Figure \ref{fig:PieceModifications}b. Finally, the original pre-designed rook was modified as depicted on Figure \ref{fig:PieceModifications}c. The final 3D chess board and pieces are shown in Figure \ref{fig:Chess3D}.

\paragraph{Blender rendering}
\label{paragraph:blender_rendering}

After texturing and modeling each of the chess pieces, a Python script was written to automate the rendering process. This script randomly assigns chess pieces to the square positions, ranging from A1 to H8. Each chess piece is randomly placed with varying degrees of centering and rotation within its assigned square. Additionally, the generated images include a proportion of each chess piece that mimics a realistic chess position, with every image incorporating both white and black kings. This guarantees a comparable class distribution to that of the real-life dataset. Further, because a continuation of this work could include using the rules of chess to apply constraints to the predictions of the model, for instance the impossibility of having more than 2 kings on the board, the generated positions were constrained to satisfy every chess rule. Finally, it is to be noted that three distinct lighting scenes observed in the original dataset were replicated within Blender. Prior to rendering, the Python \cite{python} script selects a lighting scene by randomly adding or removing light objects, thereby simulating the varying lighting conditions. All images were rendered from a camera object placed at the top of the board in order to maintain a consistent top-view perspective. Finally, a crop was applied during the rendering process to remove the rim of the board and exclusively display the 64 squares. A total of 4500 images were generated using this Python script.

\paragraph{Cropping individual squares}

Following the generation of the chessboard images, an external Python script was defined to divide these images into individual squares. Given that all images were rendered from a stationary camera perspective in Blender, the coordinates and dimensions of each square (A1 through H8) remained consistent across all examples. As such, a representative image was divided into the 64 squares by employing the Grid tool in Adobe Photoshop \cite{adobephotoshop}.

Each square was then resized to be 100x100 pixels and saved in a folder dedicated to its parent position, along with a global label array. Empty squares were assigned the label 0. Additionally, the encoding used to represent the different pieces can be found in Table \ref{Tab:labelnotation}. 

\begin{center}
     \begin{table}[!ht]
            \centering
                \begin{tabular}{ c | c | c | c | c | c | c }
                
                     Color &  Pawn & Knight & Bishop & Rook & Queen & King \\ [0.5ex] 
                \hline
                      White  & 1 & 2 & 3 & 4 & 5 & 6 \\ [0.5ex]     
                \hline 
                      Black  & 7 & 8 & 9 & 10 & 11 & 12 \\ [0.5ex]     
                \end{tabular}
            \caption{Table showing the conversion for converting chess labels to numerical values.}
        \label{Tab:labelnotation}
        \end{table}
    \end{center}

\subsection{Pre-processing}
\subsubsection{Target data}
\label{subsubsection:Target_preprocessing}

\paragraph{Contextualizing the pre-processing phase}
Ideally, the final pipeline would include pre-processing the full, unprocessed, target domain image in real-time before passing the individual squares to the trained model. However, using this approach during model training has multiple downsides, such as not being able to reduce the number of empty squares beforehand, as well as, more importantly, the much slower training time due to having to detect the board and crop the individual squares for every image of every batch as training progresses. Therefore, the pre-processing pipeline was run on every single target domain image beforehand, which simplified training the model significantly. That being said, the reader should keep in mind that in a realistic testing situation the pre-processing would be run as part of the full pipeline, and the outputs of this phase would be passed to the model in real-time.

\paragraph{Overview}

\begin{figure}[!ht]
\centering
\includegraphics[width=\textwidth]{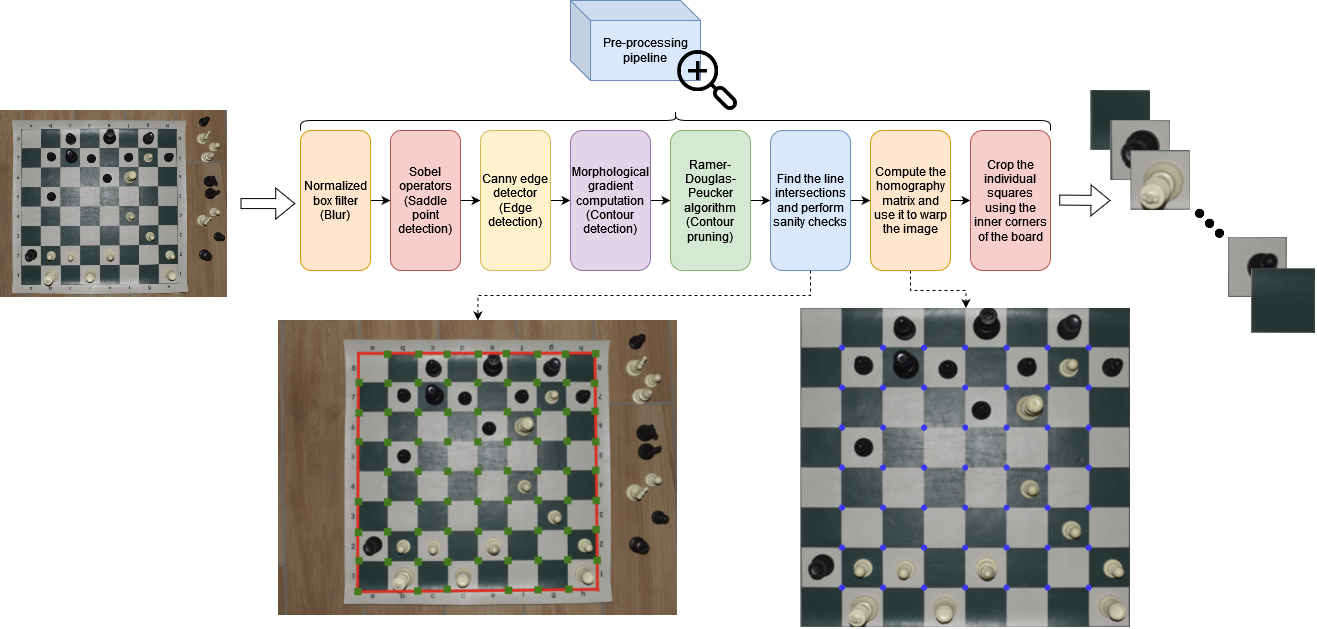}
\caption{Full overview of the pre-processing pipeline}
\label{fig:preprocessing_overview}
\end{figure}

The pre-processing phase consisted of applying a number of filters to each input target domain image in order to sequentially detect the board, warp it to the top-view, and crop the individual squares. The algorithm used was inspired by \cite{ChessRayVision}, and uses a number of filters to perform the board detection and warping, as can be shown in the overview of Figure \ref{fig:preprocessing_overview}. Please note that the computer vision part was not the main focus of this paper, and thus the mathematical details of filters will be omitted for conciseness.

The open-sourced library used to implement the algorithm can be found in \cite{ComputerVisionLibrary}. We would like to thank the authors for their contributions.

\paragraph{Detecting the board}

The board was detected using a combination of Sobel operators \cite{SobelPaper} to find the saddle points, Canny edge detection \cite{CannyPaper} to find the edges, and a mix of morphological gradient computations \cite{MorphGradPaper}, as well as the Ramer-Douglas-Peucker algorithm \cite{DouglasRamerPeuckerPaper} to detect and prune the contours, only keeping the straight lines of the image. The intersections of the straight lines are then used to find the reference points of the board, where a reference point is either defined as one of the board's corners (4 outermost points), or as the intersection between green and white squares. The outer contour of the board can then be located by linking the reference points that are located around the outer edges of the board.

\paragraph{Warping the board and cropping the squares}

Using the 4 outer corners located using the method outlined in the previous section, we can compute a 3x3 homography matrix \cite{HomographyPaper} which will help in warping the board to the top view. For a warped image of size $H \times W$, we then find  $L = min(\frac{H}{8},\frac{W}{8})$, and subsequently crop the board into 64 squares of side length $L$. The reason behind using a fixed value instead of utilizing the inner warped reference points of the board is that this approach was simpler and led to the same visual results, while also yielding squares of equal size. These squares were then resized to be 100x100 pixels to match the generated source domain data.

\paragraph{Partial Invariance to camera angle}
\label{paragraph:angle_invariance}

\begin{figure}[!ht]
\centering
\includegraphics[scale=0.2]{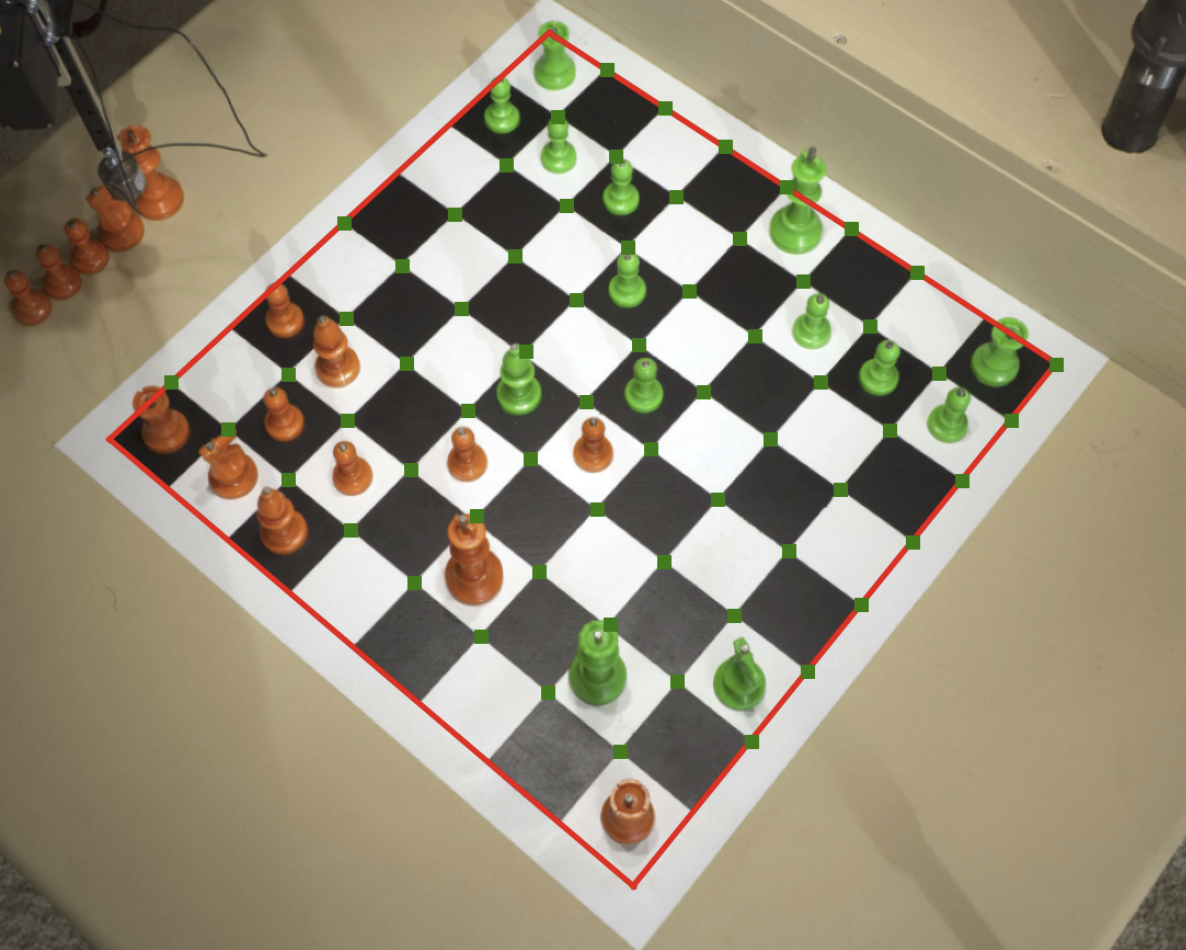}
\caption{Board detection on a side view image}
\label{fig:side_view_ex}
\end{figure}

It is important to highlight that the approach used for detecting and cropping the board means that, as long as the top of the pieces is visible enough, any image can be warped into full top-view angle and subsequently have its squares passed to the model for predictions. This applies even if the original camera angle is not entirely top-view, or even if the board is rotated, as shown in Figure \ref{fig:side_view_ex}.

\begin{figure}[!ht]
\centering
\includegraphics[scale=0.2]{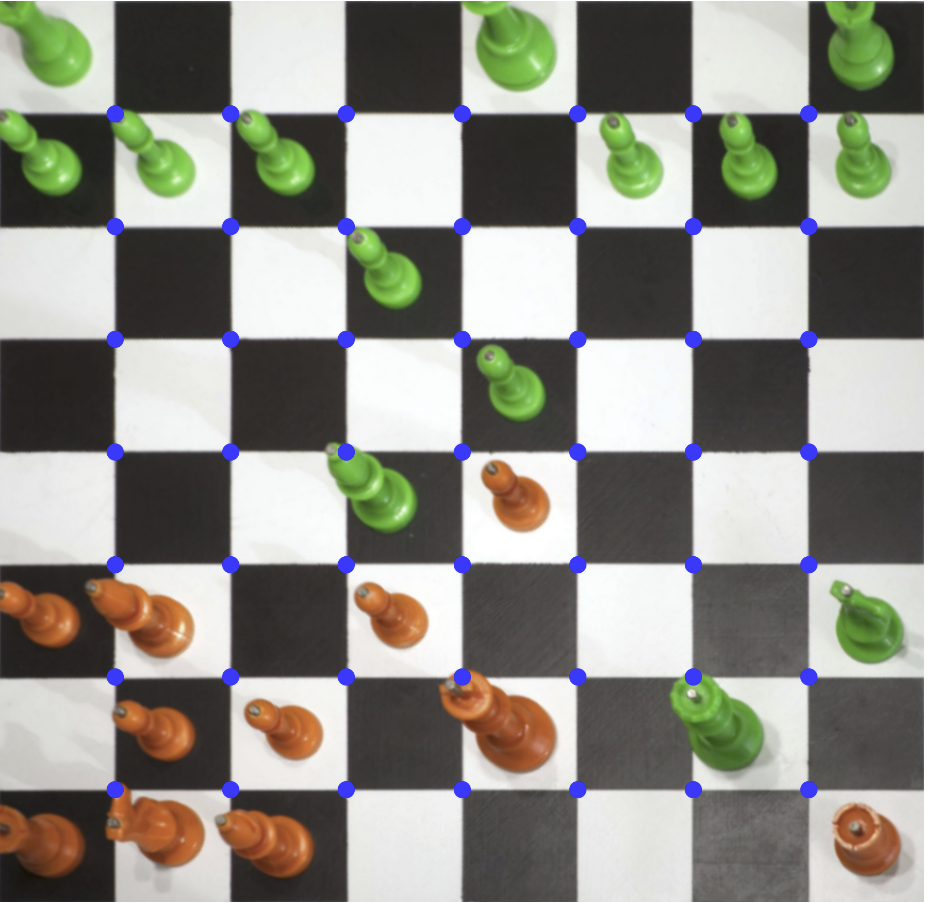}
\caption{Board warping on a side view image}
\label{fig:side_view_warped_ex}
\end{figure}

However, the downside of this approach is that, as the angle approaches the side-view, the pieces that are on the opposite side of the board relative to the camera get cut in half by the warping algorithm. Even for the example in Figure \ref{fig:side_view_ex}, there will result a partial cut of the green king as well as the two green rooks, as shown in the warped result in Figure \ref{fig:side_view_warped_ex}.

This issue will be further worsened by the square cropping algorithm, because some pieces such as the green bishop now span two squares. Taken together, these issues are the reason why the current project utilized images with a top-view camera angle. That said, an interesting continuation of this project would be to expand to other views by modifying the warping and cropping algorithms.

\subsubsection{Dataset distributions}

\begin{figure}[!ht]
    \centering
    \makebox[\textwidth][c]{%
    \begin{minipage}{2\textwidth}
        \centering
        \subfloat[Target dataset]{\includegraphics[scale=0.6]{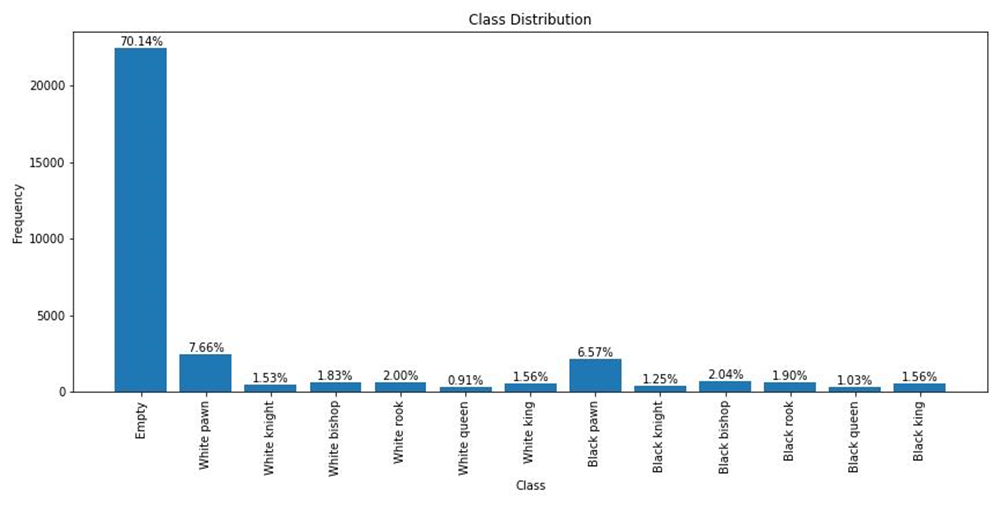}}

        \qquad

        \subfloat[Source dataset]{\includegraphics[scale=0.6]{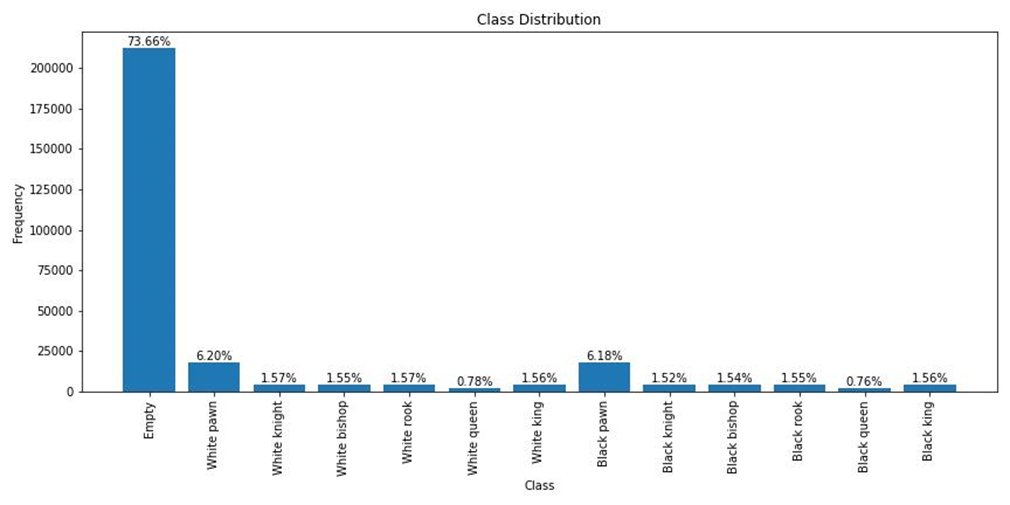}}
    \end{minipage}}
    \caption{Class distributions of the source and target datasets}
    \label{fig:NonFinalDistributions}
\end{figure}

The histograms in Figure \ref{fig:NonFinalDistributions} display the distributions of both the target data after its pre-processing and the source data after its generation. It is clear that the class distributions are imbalanced due to a high percentage of empty squares, accounting for approximately 70\% of both datasets. Additionally, an important item to highlight is that the distribution of individual chess pieces is remarkably similar across distributions, underlining the accuracy of the synthetic data generation.

Both datasets were then split into training, testing and validation sets after being randomly shuffled with the following proportions: 80\% for the training set, and 10\% for both the validation and testing sets. 

\subsubsection{Oversampling}

In order to further resolve the challenge of class imbalances in the datasets, the minority classes were oversampled. Indeed, class imbalance can negatively impact the performance of models, as they may struggle to effectively learn from under-represented classes. To address this issue, the "WeightedRandomSampler" provided by PyTorch \cite{torch} was leveraged during data loading. This sampler oversamples the minority classes, ensuring that each batch used for training received a balanced representation of samples from all 13 classes.

\subsubsection{Data augmentation}

Because oversampling entails that some training examples will be used much more frequently than others, and to further improve the model's ability to handle diverse lighting and orientation conditions, additional data augmentation techniques were applied to both datasets. This was achieved by utilizing the "transforms" module provided by PyTorch \cite{torch} during the data loading process. The applied transformations included random horizontal and vertical flips, as well as adjustments to the brightness of the image following a uniform distribution. The aim of the augmentations is to help the model in recognizing chess pieces under varying lighting conditions and orientations, thereby increasing its robustness.

\subsubsection{Data loading}

Due to the heavy size of the images, loading the full datasets into memory at once was not feasible because it required more RAM than the available computing devices could offer. To overcome this issue, a lazy data loading approach was implemented, whereby instead of loading the whole dataset at a time, the individual minibatches of images were loaded from the hard disk drive (HDD) as needed, and were subsequently discarded after the training iteration. This approach ensured that the training process could be initiated without concern for memory limitations.

\subsection{Prediction post-processing}

\begin{figure}[!ht]
\centering
\includegraphics[width=\textwidth]{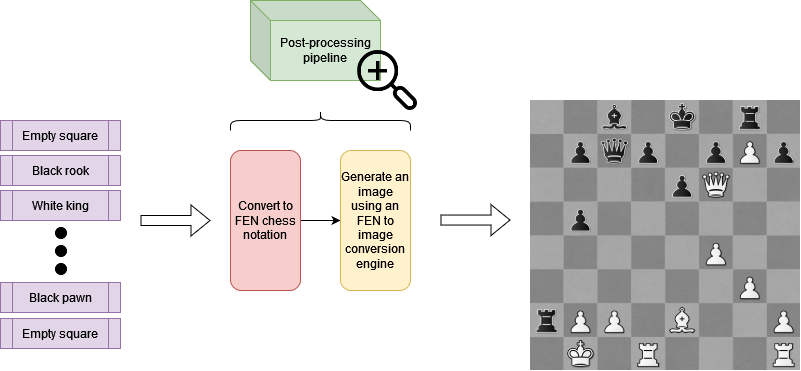}
\caption{Overview of the post-processing pipeline}
\label{fig:post_processing}
\end{figure}

During testing, the 64 individual labels predicted by the model for a given chess position are passed to a post-processing pipeline which generates the Forsyth–Edwards Notation \cite{FENWiki} of the position, which is just the standard notation used to efficiently represent the contents of a chess position. The generated string can then be used to synthesize a 2D representation of the position using open-source software such as the Fen2Image service offered by ChessVision \cite{ChessVisionFen2Image}. This process is summarized in Figure \ref{fig:post_processing}.

\subsection{High-level pipeline overview}

\begin{figure}[!ht]
\centering
\includegraphics[width=\textwidth]{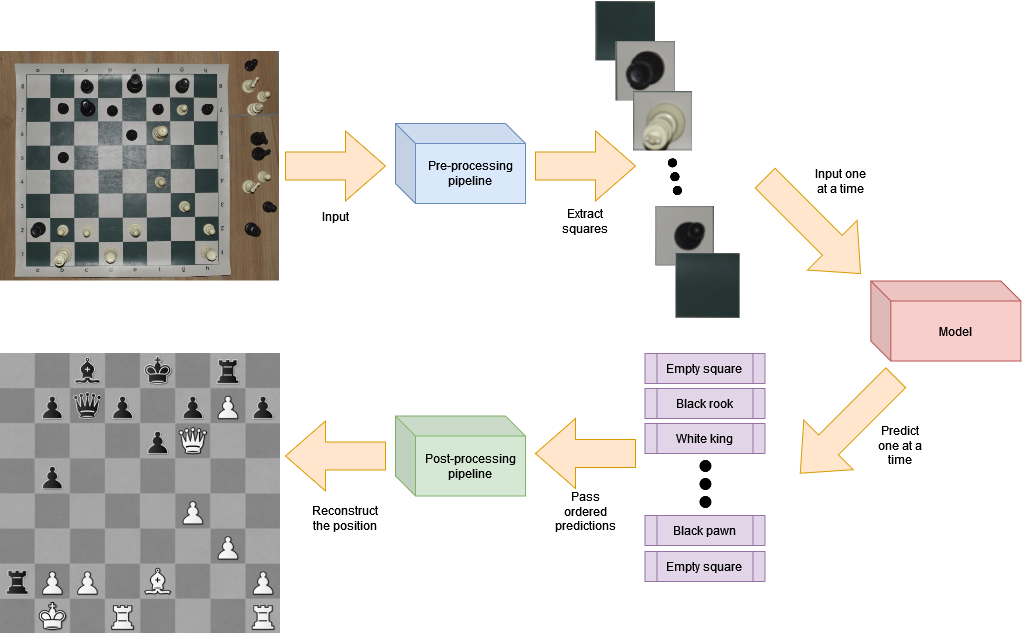}
\caption{Overview of the full pipeline}
\label{fig:full_pipeline_overview}
\end{figure}

As mentioned, although model training, validation, and testing were carried out on individual chessboard squares, the pre-processing and post-processing pipelines were designed to be a part of an end-to-end pipeline. The idea is to allow users to upload top-view, uncropped, and unprocessed photographs of their chessboards into the full pipeline and obtain a reconstruction of their chess position from it. Considering the DL model as a black box for now, an overview of the full pipeline is summarized in Figure \ref{fig:full_pipeline_overview}.

\subsection{Model}
\label{Models}

\subsubsection{Proposed approach: DANN}

\paragraph{Architecture}
\label{DANN_Arch}

\textit{1. Overview}

\begin{figure}[!ht]
\centering
\includegraphics[width=\textwidth]{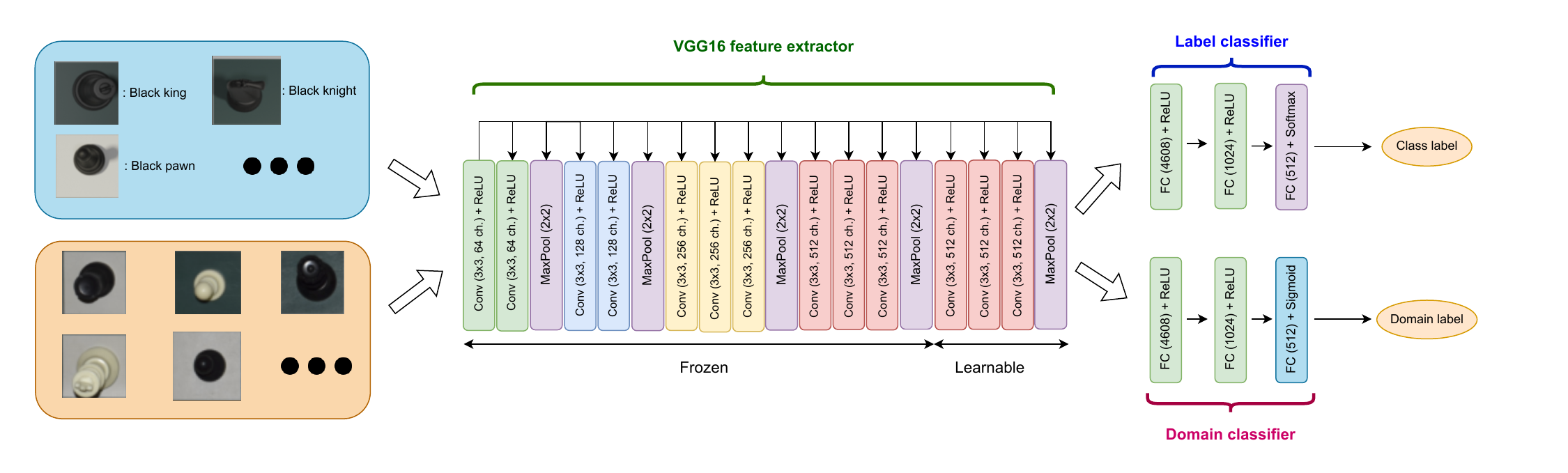}
\caption{Complete DANN architecture}
\label{fig:DANN_complete}
\end{figure}

The proposed architecture is based on a 2015 paper from Ganin et Al. \cite{DANNPaper}. At a high-level, it consists firstly of a feature extractor, which is a neural network that outputs a hidden representation of the input image. The output code is then fed to both a classifier which outputs the label of the piece on the input square, and a domain discriminator that attempts to predict whether the input is from the source or target domain. The key idea here is that the feature extractor and domain discriminator are trained following an adversarial approach. The full architecture is presented in Figure \ref{fig:DANN_complete}. Throughout subsequent sections, we will progressively explain each portion.

\textit{2. Feature extractor}

\begin{figure}[!ht]
\centering
\includegraphics[width=\textwidth]{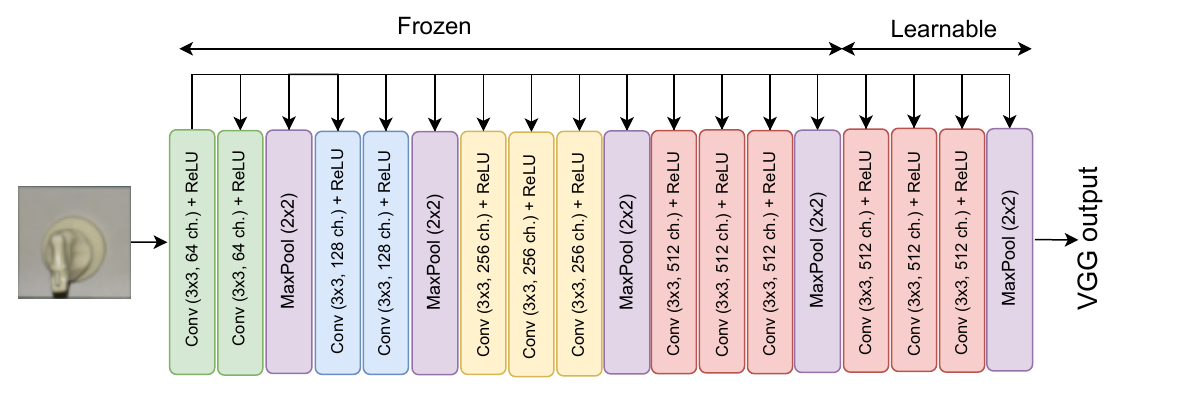}
\caption{Feature extractor architecture}
\label{fig:VGG16_DANN}
\end{figure}

The feature extractor was based on the VGG16 convolutional neural network (CNN) architecture popularized by Simonyan \& Zisserman's seminal 2014 paper \cite{VGG16_OG}. Further, the VGG16 network that was used was pre-trained on the ImageNet dataset \cite{ImageNet}. The model architecture is summarized by Figure \ref{fig:VGG16_DANN}.

During training, all but the last 3 convolutional layers were frozen, meaning that only the latter were fine-tuned.

\textit{3. Classifier}

\begin{figure}[!ht]
\centering
\includegraphics[width=\textwidth]{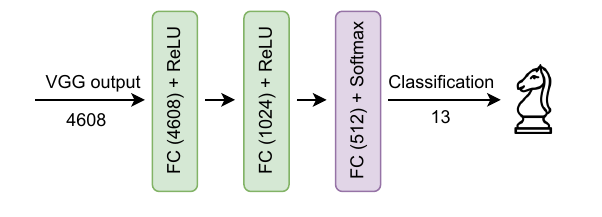}
\caption{Classifier architecture}
\label{fig:Classifier_Base}
\end{figure}

The classifier portion of the DANN consists of three fully connected linear layers using ReLU activation functions throughout. The final output is then passed to a softmax function which leads to probability predictions, as shown in Figure \ref{fig:Classifier_Base}).

\textit{4. Domain discriminator}

\begin{figure}[!ht]
\centering
\includegraphics[width=\textwidth]{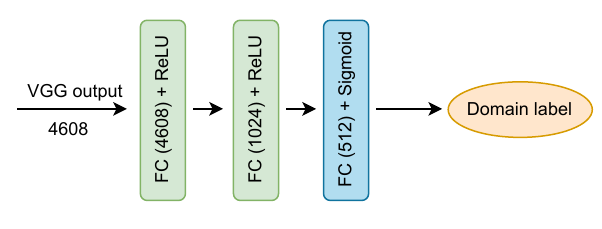}
\caption{Domain discriminator architecture}
\label{fig:DANN_Discriminator}
\end{figure}

The domain discriminator receives all outputs from the feature extractor and passes them through three fully connected linear layers with progressively decreasing widths, using ReLU activation functions. The outputs of the final layer are then passed through a sigmoid activation function for a binary prediction (see Figure \ref{fig:DANN_Discriminator}).

% The target domain data contained a total of 14,110 samples consisting of individual chess squares parsed from 500 real-life photographs of chessboard positions. The data was divided into a training split consisting 11,288 samples, a validation split consisting of 1,411 samples and a test split consisting of 1,411 samples. The source domain data contained a total of 103,533 samples consisting of individual chess squares parsed from our 4,500 images of virtual chessboard positions. The data was divided into a training split consisting of 89,226 samples, a validation split consisting of 11,153 samples and a test split consisting of 11,154 samples. 

% During training, minibatches consisting of a 50/50 split of source and target domain samples is passed to the model. The batch size was established at 100 for both training and validation splits, and the training process lasted for 30 epochs. Epoch length was determined by the number of target domain examples in the training split (11,288). Since the source domain contained a larger number of training examples (89,266), random subsets were selected to match the number of target domain examples in each epoch. Model performance was evaluated on the validation set every 10 iterations on two consecutive batches whose results were averaged. The final validation run was assessed on 20 consecutive batches whose results were also averaged and displayed in Section \ref{subsubsection:DANN_results}.

\paragraph{Loss functions}

\textit{1. Classifier loss}

The most straightforward and widely used approach to assessing multi-class classification performance is categorical cross-entropy loss \cite{CrossEntropyPaper}:

\[
    \text{CE} = -\log(p_t)
\]

where $p_t$ is the predicted probability for the true class.
The problem with this function for chess piece classification is that it assigns equal weight to all classes in the total loss calculation. However, some classes are clearly much easier to predict than others, which could give the impression that the classifier is doing well when it is in fact performing poorly. In such cases, a generalized version of cross-entropy called Focal Loss \cite{FocalLoss} can be used:

\[
    \text{FL} = -(1-p_t)^{\gamma}\log(p_t)
\]

As $p_t$ becomes larger, meaning that the model is finding it easier to classify the correct class, $(1-p_t)$ decreases and less weight is assigned to that class. As such, Focal loss assigns less weight to classes that are easier to classify and more weight to classes that are more difficult to classify. Further, as the hyperparameter $\gamma$ increases, the loss function down-weights classes that are easier to classify incrementally more. This is best illustrated with Figure \ref{fig:Gamma_plot} which is taken from \cite{FocalLoss}: 

\begin{figure}[!ht]
\centering
\includegraphics[width=0.8\textwidth]{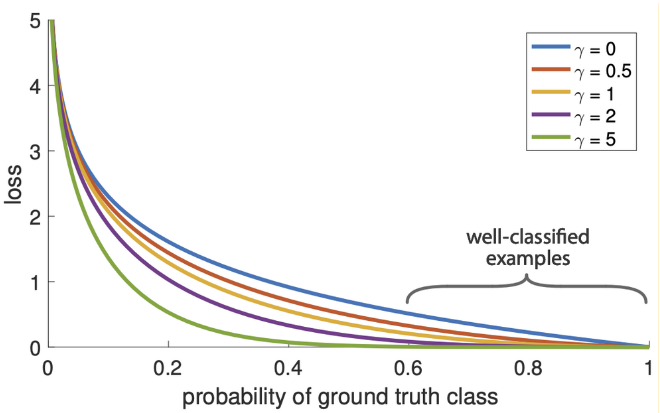}
\caption{Gamma hyperparameter and its effect on Focal loss}
\label{fig:Gamma_plot}
\end{figure}

It can be observed that, for regular cross-entropy, which is just a special case of Focal loss where $\gamma = 0$, an example predicted correctly with probability 0.6 is assigned a loss value of approximately 1, while for $\gamma = 5$ the loss value is practically 0.

\textit{2. Discriminator loss}

The discriminator's error was calculated using the Binary Cross Entropy (BCE) loss:

\[
    \text{BCE}(p, y) = -y\log(p) - (1-y)\log(1-p)
\]

where $p$ is the predicted probability for the positive class and $y$ is the true label.

\paragraph{Training overview}
\label{paragraph:DANN_training}

\textit{1. Forward pass}

During a forward pass, minibatches consisting of a 50/50 split of source and target domain samples are passed to the model in succession. After passing all examples through the feature extractor, the resulting hidden representations are passed through both the classifier and the domain discriminator for the source domain, while the codes for the target domain data are only passed to the domain discriminator, since no labels are available for that portion of the minibatch.

\textit{2. Backward pass}

\begin{figure}[!ht]
\centering
\includegraphics[width=\textwidth] 
    {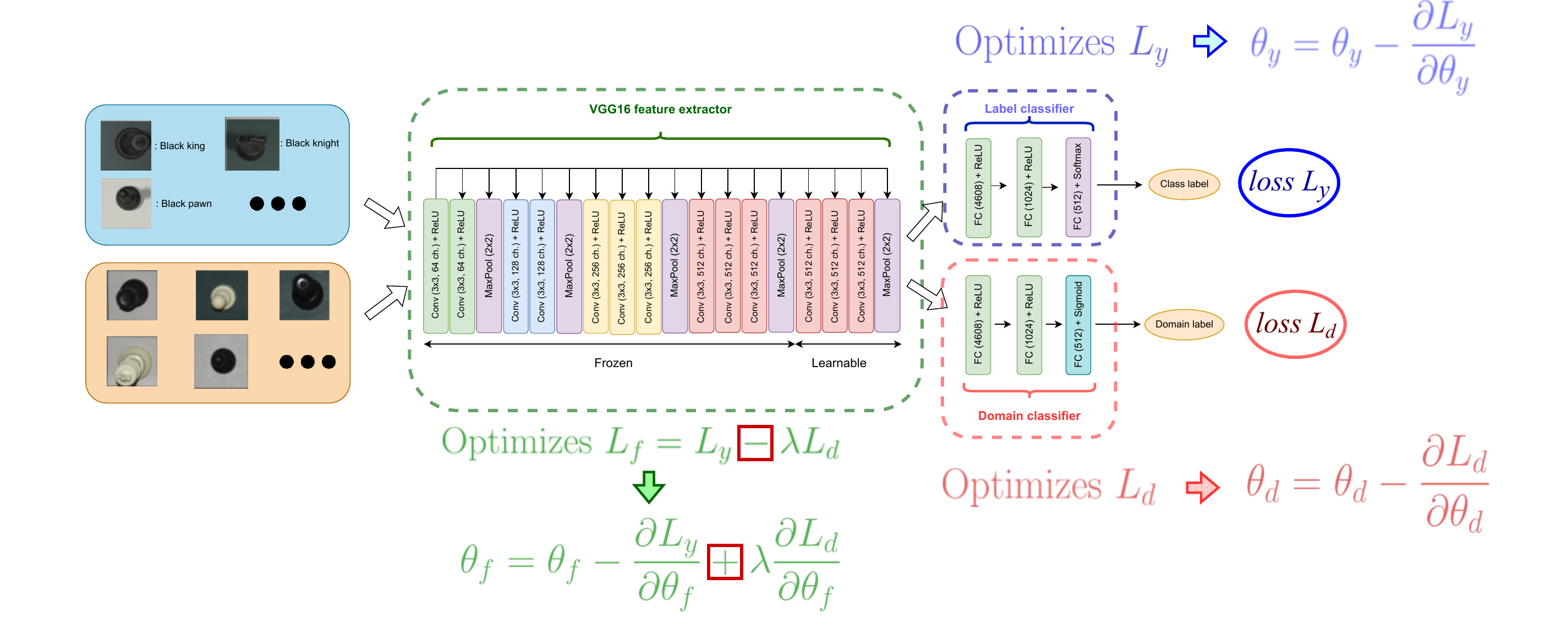}
\caption{Complete DANN architecture with loss functions}
\label{fig:DANN_LossFx_plot}
\end{figure}

As for any neural network trained through backpropagation, the weights of the classifier and the domain discriminator were updated to optimize the Focal and BCE losses, respectively. The domain adaptation, however, actually occurs through the adversarial training of the feature extractor and domain discriminator. Indeed, assuming $L_y$ is the classification loss, and $L_d$ is the discrimination loss, the feature extractor, instead of optimizing the sum of the two quantities, will optimize their difference:
\[
    L_f = L_y - \lambda L_d
\]
where $L_f$ refers to the loss optimized by the feature extractor, and $\lambda$ is a hyperparameter that controls the strength of the domain adaptation process. This is summarized in Figure \ref{fig:DANN_LossFx_plot}.
In simple terms, this means that the feature extractor will try to minimize the classification loss while maximizing the discrimination loss. Because the discriminator will attempt to optimize its own weights in order to minimize that same loss, it ensures that training a DANN turns into a min-max optimization problem where the feature extractor is attempting to confuse the domain discriminator, while the latter is trying to differentiate the two domains. Through evolving together, the aim is to reach an equilibrium where the feature extractor is able to confuse the domain discriminator by outputting similar representations for both domains, making the good classification accuracy on source domain images port well to the target domain.

\begin{figure}[!ht]
\centering
\includegraphics[width=\textwidth]{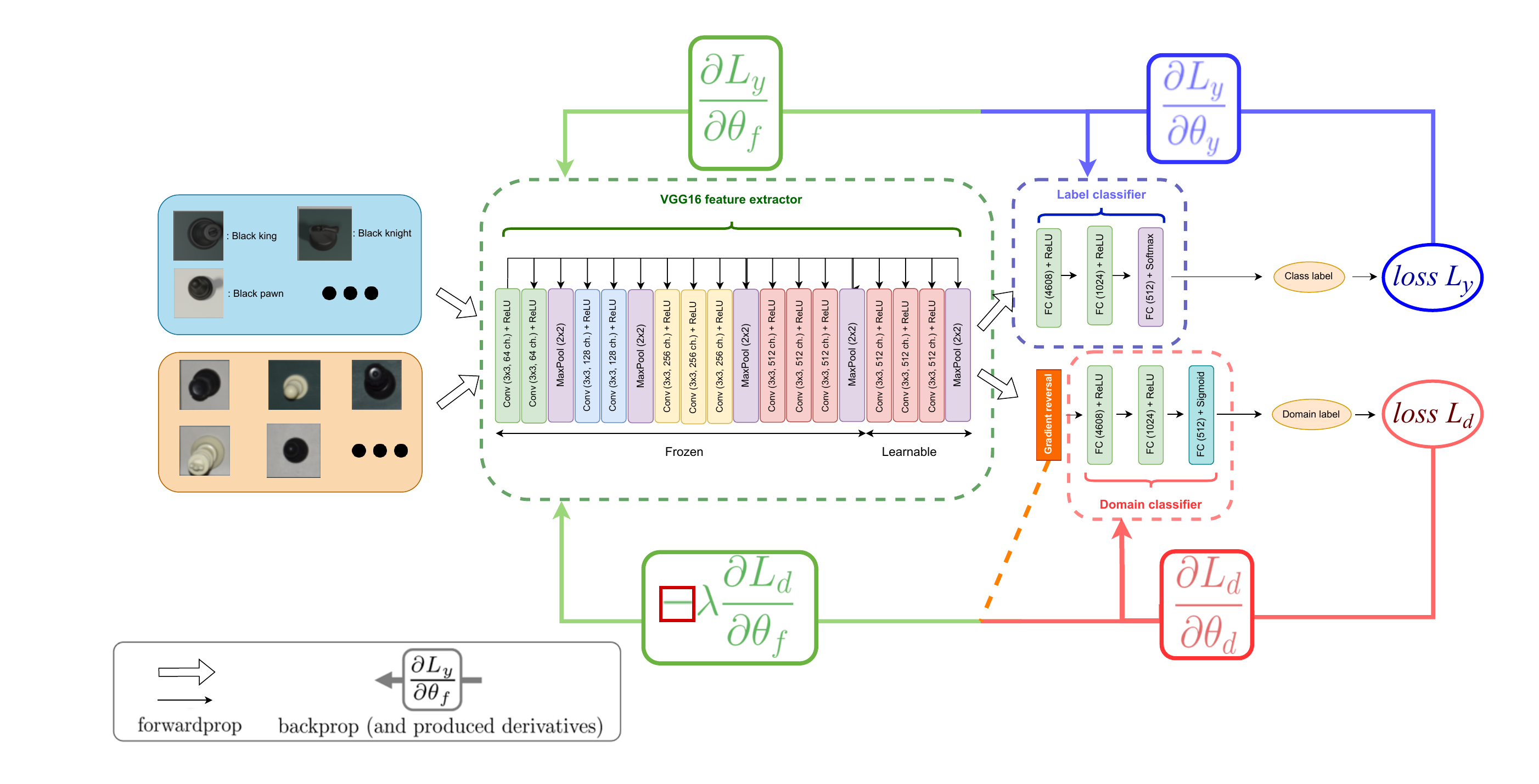}
\caption{Backpropagation phase for the DANN model}
\label{fig:DANN_GradRev}
\end{figure}

Finally, implementation of the aforementioned optimization problem was achieved through a gradient reversal layer at the start of the domain discriminator, as recommended by \cite{DANNPaper}. During the forward pass, the gradient reversal layer acts as an identity function. However, during backpropagation, it multiplies the gradient passed through it by $-\lambda$, basically reversing the gradient and applying the domain adaptation factor. This is summarized in Figure {\ref{fig:DANN_GradRev}}.

It is to be noted that a regular Stochastic Gradient Descent (SGD) optimizer with a momentum factor of 0.9 \cite{SGDPaper} was chosen, instead of a more sophisticated approach such as the Adaptive Moment Estimation (ADAM) \cite{AdamPaper}. This is because training an adversarial network is a very fragile process, and thus it is best to avoid interfering with the learning rate to facilitate isolating the potential reasons for the training not going in the expected direction.

\textit{3. Metrics tracked}

During training, the classifier and discriminator losses were tracked through epochs. In addition, the classification performance with respect to source domain data was monitored using standard classification accuracies and weighted F1 scores. Because of the oversampling applied to the data, classes were balanced, and thus standard accuracy shown below is a viable option:

\[
Accuracy = \frac{\# \text{correctly classified inputs}}{\# \text{inputs}} = \frac{TP + TN}{TP + TN + FP + FN}
\]

where TP, TN, FP, and FN are the respective numbers for true positives, true negatives, false positives, and false negatives classified.

Further, the F1 score is the harmonic mean of precision and recall, where precision is the proportion of true positive predictions among all predicted positive instances, and recall is the proportion of true positive predictions among all actual positive instances \cite{F1paper}:

\[
F1 = 2 \cdot \frac{\text{precision} \cdot \text{recall}}{\text{precision} + \text{recall}}
\]

The weighted F1 score metric additionally computes a weighted average of the F1 scores across all classes, where the weights defined below as $w_i$ represent the number of instances of class $i$, and $F1_{i}$ is the score of class $i$.

\[
F_{1}^{w} = \frac{\sum_{i=1}^{n} w_{i} \cdot F1_{i}}{\sum_{i=1}^{n} w_{i}}
\]

\paragraph{Validation}
\label{paragraph:DANN_validation}

\textit{1. Metrics tracked}

During validation, the labels that were kept for the target domain's validation set were utilized and the model’s Focal losses, Binary Cross Entropy losses, standard classification accuracies, and weighted F1 scores were tracked for both source and target domain data. This is in contrast to training, where these metrics were only tracked for the source domain.

\textit{2. Hyperparameter tuning strategy}

The two hyperparameters which were tuned during training were the learning rate t,he $\lambda$ domain adaptation factor, and the $\gamma$ focal loss factor. 

The learning rate values attempted are 0.01 and 0.02, and an SGD optimizer was used with a momentum value of 0.9.

As for the domain adaptation factor $\lambda$, the values attempted were 0.2, 0.3, as well as a dynamic range of values whose calculation was derived as follows: 

\[
{\lambda} = \frac{2}{1 + e^{-\gamma p}} - 1
\]

Where $p$ represents the linear training progress from 0 to 1, and $\gamma$ is a scaling factor fixed to a value of 10, as suggested by \cite{DANNPaper}. This dynamic range of $\lambda$ values is denoted as \textit{variablepaper} in the hyperparameter tuning tabular results in Section \ref{subsubsection:DANN_results}.

For the Focal loss function, the values of the hyperparameter $\gamma$ attempted were 2 and 5 to attempt different degrees of prioritizing the classes that are more difficult to predict.

\subsubsection{Alternative baseline 1: Base-Source model}
\label{subsubsec:Base_model_source}
\paragraph{Architecture}

\begin{figure}[!ht]
\centering
\includegraphics[width=\textwidth]{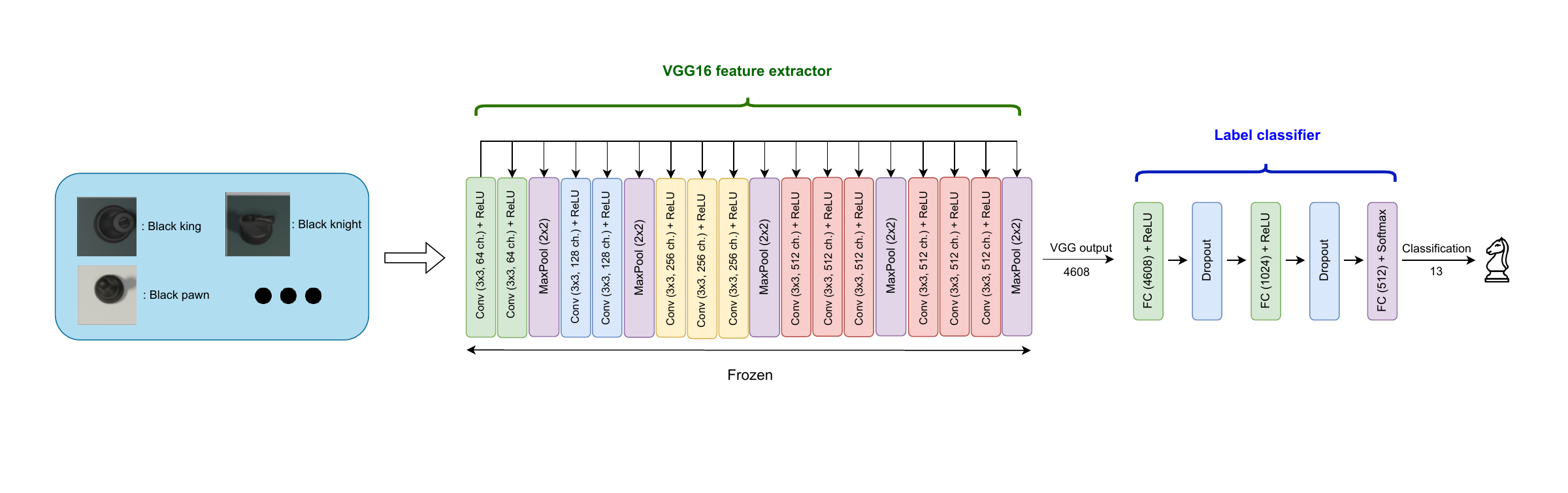}
\caption{Full architecture of the Base-Source model}
\label{fig:BaseArch}
\end{figure}

In order to benchmark the performance of the DANN, the same VGG16 architecture described in Section \ref{DANN_Arch} was utilized for the feature extractor. This was followed by three fully connected layers, interleaved with dropout layers to avoid overfitting, leading up to the classification. This VGG16 was also pre-trained on ImageNet, with the difference being that all of its layers were frozen, and only the fully connected portion was fine-tuned. This is summarized in Figure {\ref{fig:BaseArch}}.

\paragraph{Loss function}
The Base-Source model utilized the same Focal loss used to assess the classification error of the DANN model.

\paragraph{Training overview}

\textit{1. Forward and backward passes}

The training process was conducted using only the source domain data, without utilizing any domain adaptation whatsoever. The backpropagation update formulas were applied using an ADAM \cite{AdamPaper} optimizer.

\textit{2. Metrics tracked}

The metrics tracked during training of the Base-Source model are Focal loss, standard accuracy and weighted F1 scores for the source domain data.

\paragraph{Validation}
\label{paragraph:Base_validation}

\textit{1. Metrics tracked}

During validation, the same metrics used during training were tracked, but this time for both the source and target domains, since labels for the validation subset of the target data were now available.

\textit{2. Hyperparameter tuning strategy}

While fine-tuning the Base-Source model, the majority of the hyperparameters were fixed. Only the influence of the focal loss $\gamma$ values and dropout rates were assessed. The Focal loss $\gamma$ was set to both 0.2 and 0.5 sequentially so as to determine the optimal strength of the down-weighting of easy examples. Finally, dropout rates of 0.2 and 0.5 were attempted. The learning rate was fixed to be 0.001 and the batch size to 100.

\subsubsection{Alternative baseline 2: CORAL model}

\paragraph{Architecture}

The CORAL model has the same architecture as the Base-Source model, except for 3 key differences:
\begin{itemize}
    \item The dropout layers of the classifier portion are discarded.
    \item The last 3 convolutional layers of the VGG16 are made learnable.
    \item Another loss function, the correlation alignment loss (CORAL), is computed at the first and last linear layers of the classifier
\end{itemize}

\begin{figure}[!ht]
\centering
\includegraphics[width=\textwidth]{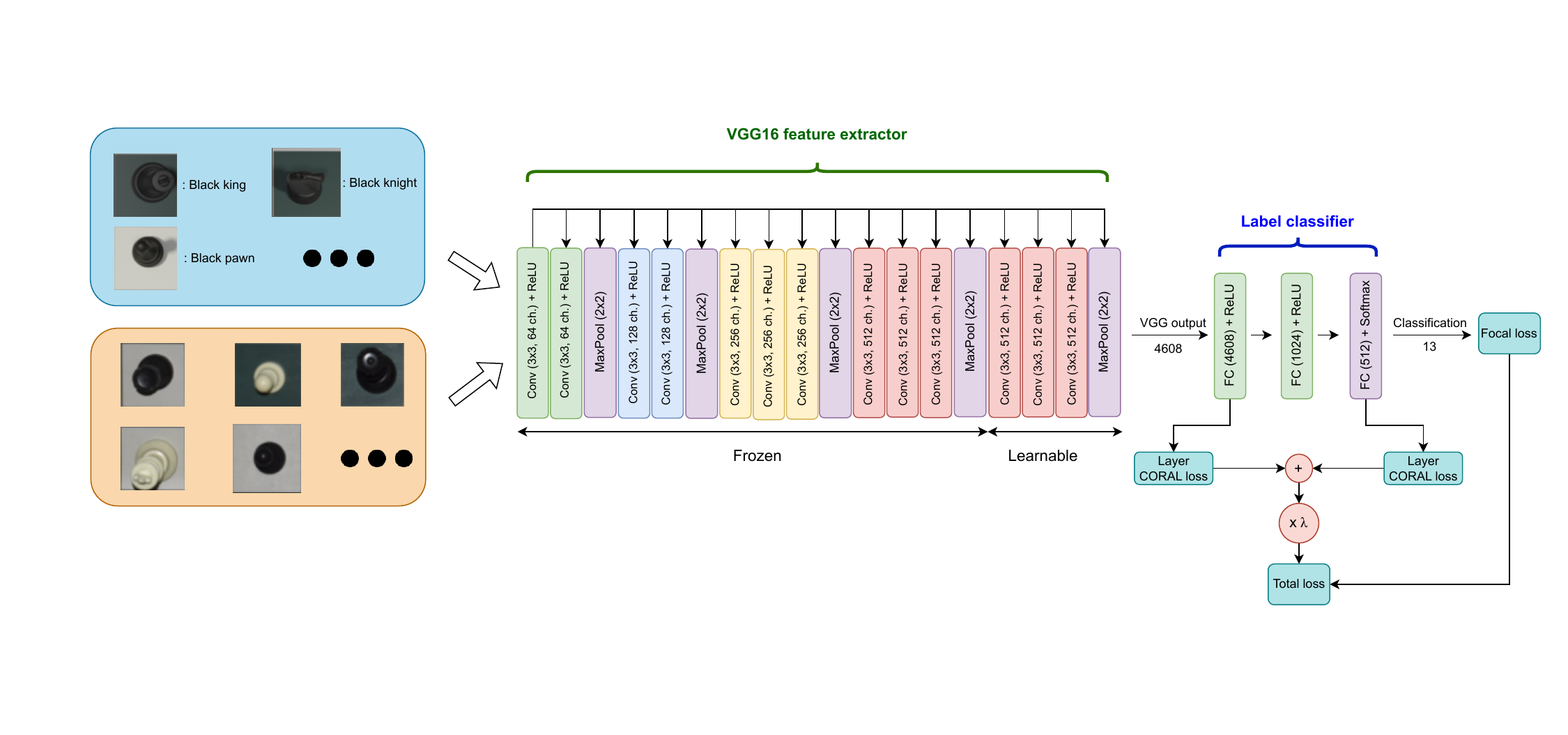}
\caption{Complete CORAL model architecture}
\label{fig:CORAL_Arch}
\end{figure}

The reason for the first change is to make the classifier portion of the CORAL model be the same as that of the DANN model, rendering the comparison equitable. The reason for the second \& third changes will become apparent in the training section. The architectural changes applied are shown in Figure {\ref{fig:CORAL_Arch}}. It is to be noted that the intuition behind $\lambda$ will become apparent in the training section.

\paragraph{Loss functions}

The loss functions utilized by the CORAL model during training included Focal loss for the label classification, as well as CORAL loss, which aims to minimize the domain shift between two distributions by aligning the second-order statistics of their features \cite{CORALPaper}. It computes the covariance matrix of the source and target features and then minimizes the Frobenius norm between the difference of the two covariance matrices. This way, the correlation between the features is preserved and domain shift is reduced. It can be obtained using:

\[
{L}_{\text{CORAL}} = \frac{1}{4d^2}\|C_s - C_t\|_F^2
\]

Where $d$ refers to the dimensionality of the feature space, $4d^2$ is a scalar constant that scales the loss function to make the magnitude more interpretable, $C_S$ and $C_T$ are the covariance matrices of the source and target data, and $|\cdot|_F^2$ denotes the squared Frobenius norm. The covariance matrices of the source and target data are specifically given by:

\[
C_S = \frac{1}{n_S-1}(X_S - \bar{X}_S)(X_S - \bar{X}_S)^T
\]

\[
C_T = \frac{1}{n_T-1}(X_T - \bar{X}_T)(X_T - \bar{X}_T)^T
\]

Where $n_S$ and $n_T$ denote the number of samples in the source and target domain, $X_S$ and $X_T$ denote the data matrices for the source and target domain, and the terms $\bar{X}_S$ and $\bar{X}_T$ represent the column-wise mean of $X_S$ and $X_T$, respectively.

\paragraph{Training overview}

\textit{1. Forward and backward passes}

During a forward pass, minibatches consisting of a 50/50 split of both domains were passed to the model. The source domain examples are the only portion that reach the classification stage, while both domains are equally used to compute the CORAL losses at the first and last linear layers of the classifier. The total loss of the model can then be computed as:

\[
TL = FL + \lambda ({L}_{\text{CORAL}})
\]

where $FL$ represents the Focal loss, and where increasing the value of $\lambda$ assigns more weight to the CORAL loss value, as was the case for the domain adaptation factor of the DANN model. The challenge is to find a value of $\lambda$ that is large enough so that sufficient domain adaptation occurs in order to predict the target domain accurately, but not too large so that the classifier is neglected by the optimizer, leading to poor accuracies in both domains. 

The hope is that, through minimizing the CORAL loss, the hidden representations of the two domains will become similar enough so that predictions that are accurate for the source domain will work comparably well for the target domain.

After the losses are computed, backpropagation is performed using an ADAM optimizer, just as for the Base-Source model.

\textit{2. Metrics tracked}

During training, various metrics were tracked, including Focal loss, standard accuracy, and weighted F1 scores, all for the source domain data. Further, the model's training performance was also assessed based on its total CORAL loss (Sum of the two CORAL losses computed), as well as the total loss, as outlined in the preceding subsection.

\paragraph{Validation}
\label{paragraph:CORAL_validation}

\textit{1. Metrics tracked}

During validation, Focal losses, standard classification accuracies, and weighted F1 scores were monitored for both source and target domain data, making use of the labeled validation subset of the target data. The model's validation performance was also assessed based on its aggregated layer CORAL loss as well as its total loss, with the formula for the latter using the Focal loss of the source domain data only (not the sum of the Focal losses of both domains) in order to stay on the same scale as the training total loss.

\textit{2. Hyperparameter tuning strategy}

The three hyperparameters which were tuned during training were the learning rate, with attempted values of 0.0005 and 0.001, the focal loss' $\gamma$ parameter for which the values of 2 and 5 were tested, as well as the domain adaptation factor $\lambda$.

The classifier was given a head start before gradually increasing the degree of domain adaptation by progressively incrementing $\lambda$ from 0 to $\lambda_{\text{max}}$ throughout epochs, using the formula:
$$\lambda = p \times \lambda_{\text{max}}$$
Where $p$ indicates the linear progress of the training going from 0 to 1, and the $\lambda_{\text{max}}$ values of 0.01, 0.1, 1, 10 and 100 were attempted.

As indicated in the next section, many combinations of only these three hyperparameters were tested because their interplay led to drastically different performance rates (see Section \ref{CORAL_results}).

Meanwhile, the batch size was fixed to 100 and the number of epochs to 30, giving the model enough time to converge

\subsubsection{Alternative baseline 3: Base-Target model}
\label{subsubsec:Base_model_target}
This is the same model as the first alternative baseline in Section \ref{subsubsec:Base_model_source} except that it is trained directly on the labeled target domain data. However, since the whole idea of this project is based on the premise that the target domain labels are not available, this baseline is only included as a comparison point, giving the reader an idea of what an upper bound on the maximum possible domain adaptation performance achievable. All the hyperparameter combinations attempted are the same.

\section{Results}
\label{sec:Results}

\subsection{DANN}

\subsubsection{Hyperparameter tuning results}
\label{subsubsection:DANN_results}

The hyperparameter tuning for the DANN model was done with the following fixed parameters: $Batch\;Size = 100$ ; $Number\;of\;epochs = 30$. The best validation accuracy value on the target dataset was 84.1\%, obtained for $ \gamma \text{ (Focal loss)} = 5$; $\lambda \text{ (DA factor)} = 0.3$ and $Learning\;rate = 0.02$, as shown in Table \ref{tab:DANNHP}.

\begin{table}[htbp]
\centering
\begin{tabular}{cccc}
\toprule
\textbf{$\gamma$ (Focal Loss)} & \textbf{Learning Rate} & \textbf{$\lambda$ (DA factor)} & \textbf{Best Validation Accuracy} \\ \midrule
2 & 0.02 & 0.2 & 82.3\% \\
2 & 0.02 & variable & 82.4\% \\
2 & 0.02 & 0.3 & 84.4\% \\
2 & 0.01 & 0.2 & 84.2\% \\
2 & 0.01 & variable & 82.6\% \\
2 & 0.01 & 0.3 & 78.8\% \\
5 & 0.02 & 0.2 & 84.8\% \\
5 & 0.02 & variable & 79.9\% \\
\textbf{5} & \textbf{0.02} & \textbf{0.3} & \textbf{85.0\%} \\
5 & 0.01 & 0.2 & 78.9\% \\
5 & 0.01 & variable & 77.7\% \\
5 & 0.01 & 0.3 & 76.5\% \\
\bottomrule
\end{tabular}
\caption{DANN hyperparameter tuning results}
\label{tab:DANNHP}
\end{table}

As a reminder, the \textit{variable} value for $\lambda$ refers to the suggestion by the DANN paper \cite{DANNPaper} that makes $\lambda$ increase from 0 to 1 throughout the training, as explained in section \ref{paragraph:DANN_validation}.

\subsubsection{Validation curves}
\label{subsubsection:DANN_Val_Curves}

\begin{figure}[!ht]
    \centering
    \makebox[\textwidth][c]{%
    \begin{minipage}{\textwidth}
        \centering
        \subfloat[Validation Discriminator Losses]{\includegraphics[scale=0.3]{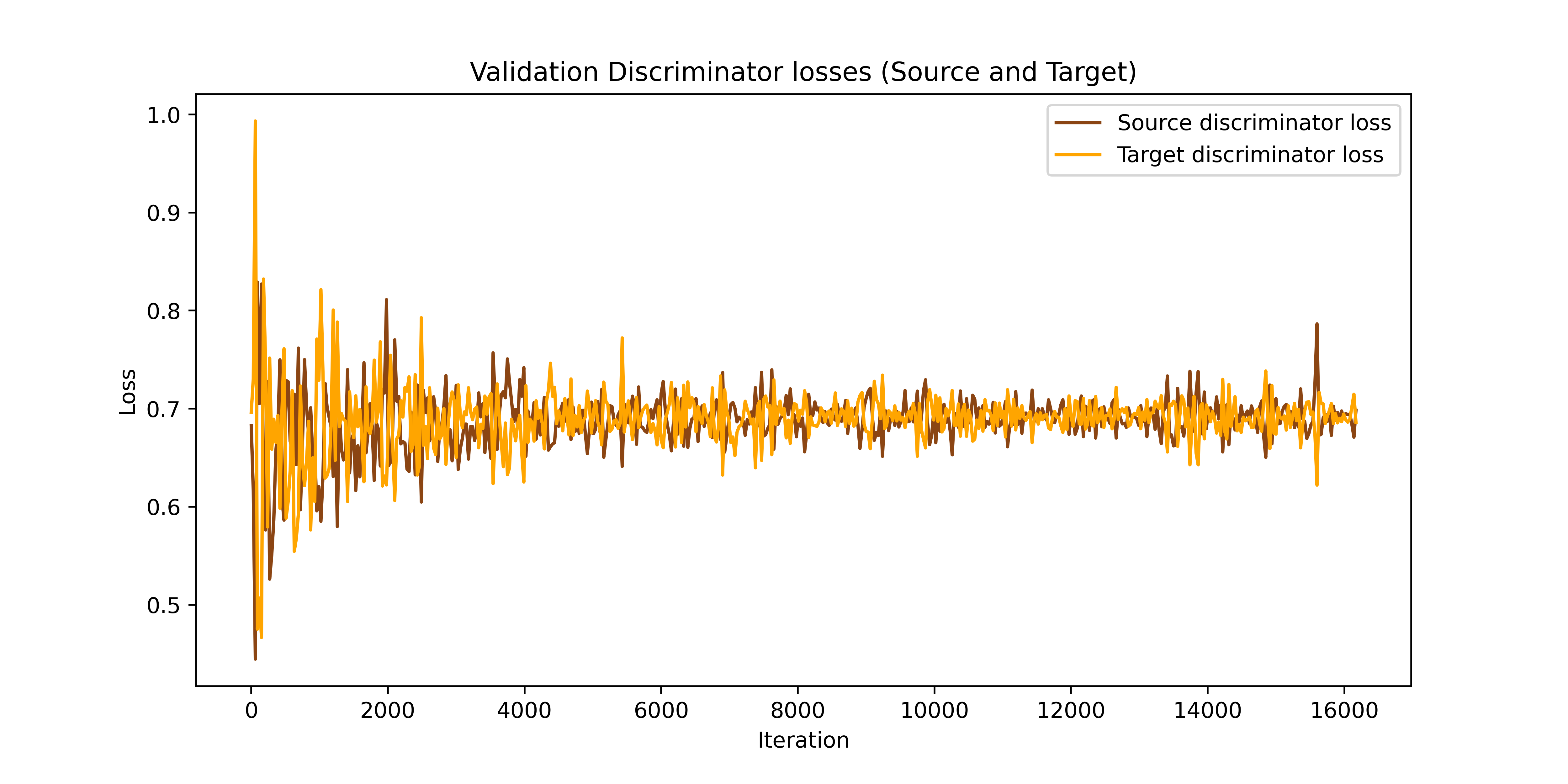}}
        \qquad
        \subfloat[Validation Classifier Losses]{\includegraphics[scale=0.3]{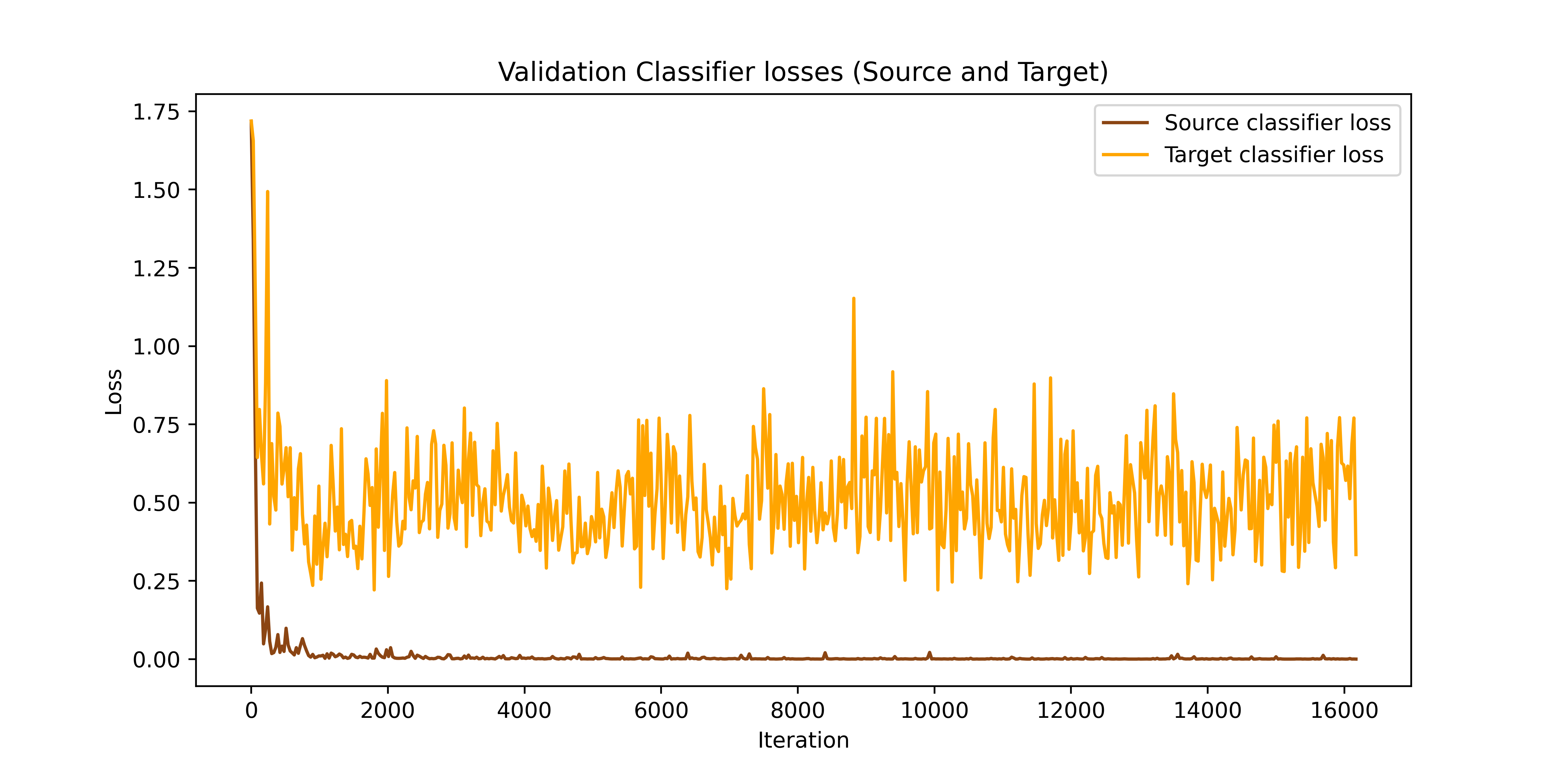}}
        \qquad
        \subfloat[Validation Accuracies]{\includegraphics[scale=0.3]{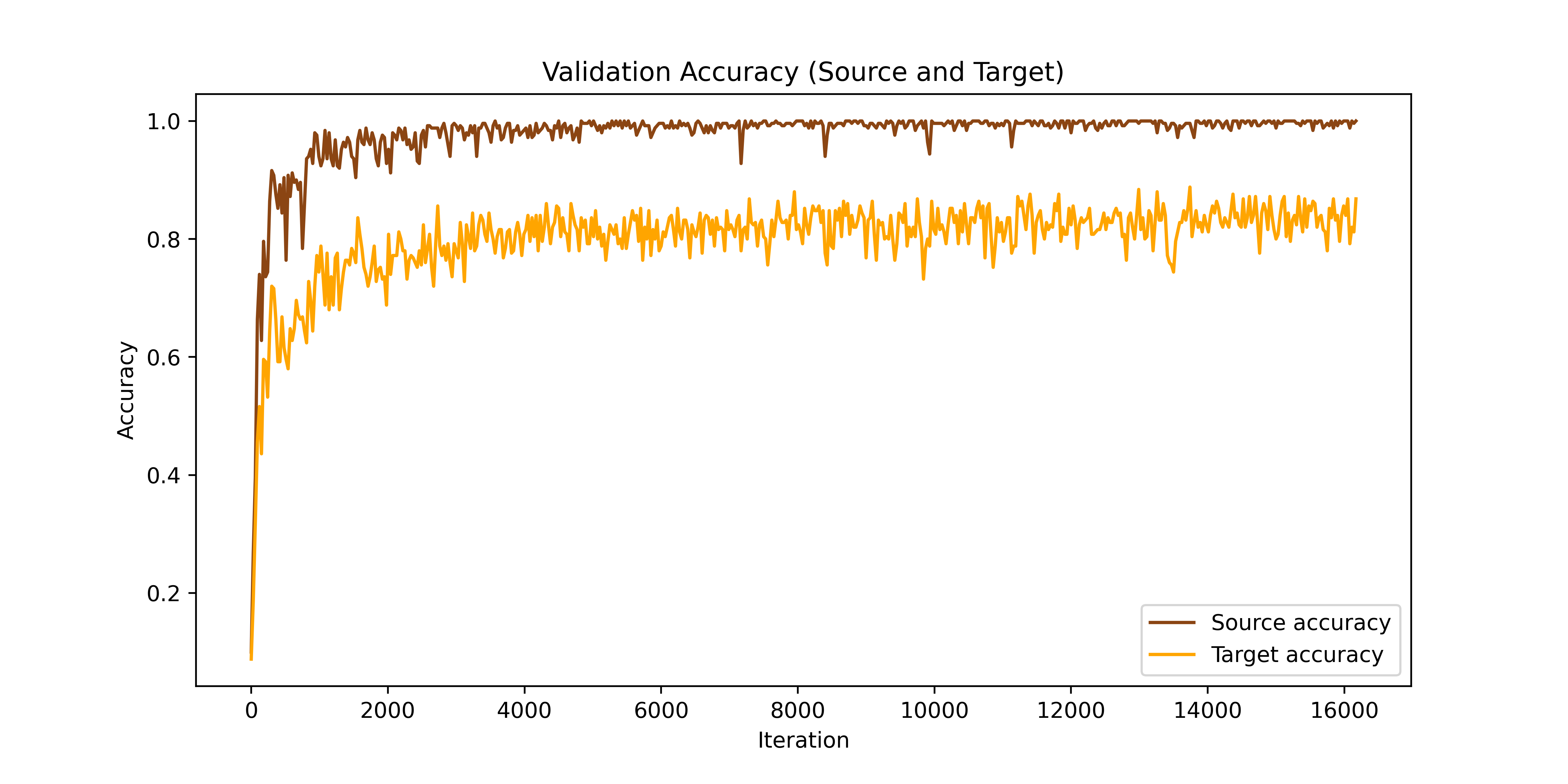}}
        \qquad
        \subfloat[Validation weighted F1-score]{\includegraphics[scale=0.3]{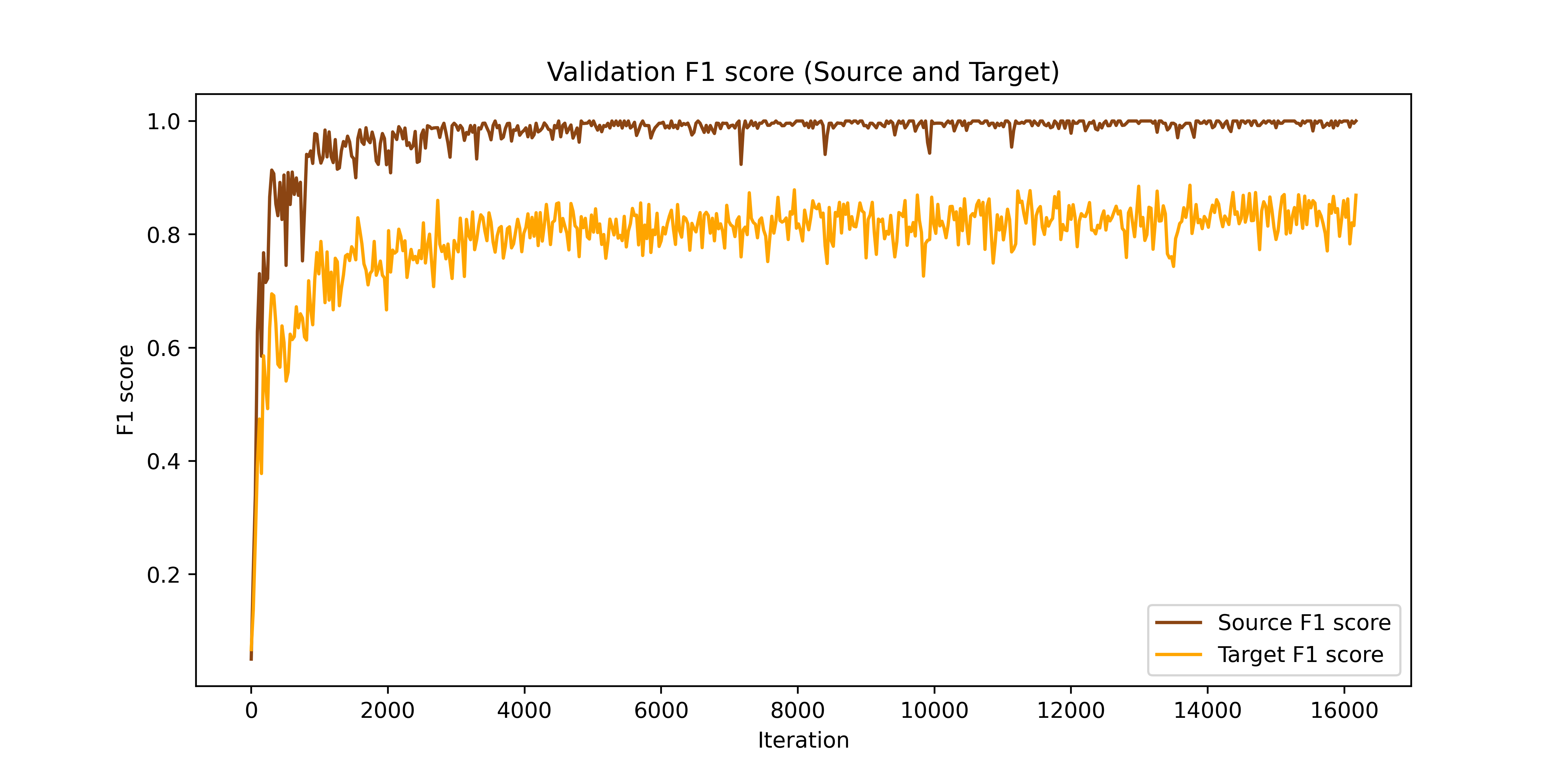}}
    \end{minipage}}
    \caption{Final DANN Model validation curves}
    \label{fig:DANNValidation}
\end{figure}

The validation metrics mentioned in Section \ref{paragraph:DANN_validation} corresponding to the hyperparameters that yielded the best validation accuracy (as shown in Table \ref{tab:DANNHP}) are depicted in Figure \ref{fig:DANNValidation}.

The discriminator loss plots of both domains in Figure \ref{fig:DANNValidation}a show oscillations that, except for the first few iterations, gradually increase, converging to around 0.7. These oscillations are a reflection of the dynamic competition between the feature extractor and the domain classifier. Indeed, while the discriminator aimed to accurately classify the domain of the input data, the feature extractor was consistently improving to generate hidden feature representations that are domain-invariant, thus making it difficult for the discriminator to accurately classify the domain. The challenge resided in striking a balance between the discriminator learning too fast for the feature extractor when $\lambda$  was too big, or the feature extractor out-learning the discriminator when $\lambda$  was too small. It is to be noted that, while the feature extractor should aim to fool the discriminator, the discriminator needs to be performing well to begin with. Otherwise, there is no point in confusing a discriminator that consists of only random weights. 

In an ideal scenario, the discriminator would become totally random despite its attempts to correct its predictions. The discriminator of the DANN model in this project converged at a BCE loss of around 0.7. Assuming that $p_t$ is the probability predicted for the true class, this means that:
$$\text{BCE} = 0.7$$
$$-\log(p_t) = 0.7$$
$$p_t = 2^{-0.7} \simeq 0.6$$
which is very close to the desired 0.5. This shows that the feature extractor nearly reached its full domain-invariance potential. On the other hand, the classifier losses for both domains depicted on Figure \ref{fig:DANNValidation}b showed a decrease over iterations. This indicates that, in addition to the feature extractor generating domain-invariant features, the classifier also effectively learned to classify properly, thereby proving that $\lambda$ was not too big. As a result, the validation accuracies as well as the weighted F1-scores for both domains shown in Figures \ref{fig:DANNValidation}c and \ref{fig:DANNValidation}d increased as iterations progressed.

\subsection{Base-Source model}

\subsubsection{Hyperparameter tuning results}
\label{subsubsection:Base_results_source}

The results of the fine-tuning of the Base-Source model are shown in Table \ref{tab:BaseSourceHP}. This hyperparameter tuning process was executed using the following fixed parameters: $Learning\;rate = 0.001$, $Batch\;Size=100$, and $Number\;of\;epochs = 5$. The best validation accuracy value on the target dataset was 61.4\%, found with $\gamma= 2$,  and $Dropout\;rate=0.2$.

\begin{table}[htbp]
\centering
\begin{tabular}{ccc}
\toprule
\textbf{$\gamma$ (Focal loss)} & \textbf{Dropout Rate} & \textbf{Best Validation Accuracy} \\ \midrule
2 & 0 & 59.7\% \\
\textbf{2} & \textbf{0.2} & \textbf{61.4\%} \\
2 & 0.5 & 56.7\% \\
5 & 0 & 56.6\% \\
5 & 0.2 & 56.9\% \\
5 & 0.5 & 58.4\% \\
\bottomrule
\end{tabular}
\caption{Base-Source hyperparameter tuning results}
\label{tab:BaseSourceHP}
\end{table}

\subsubsection{Validation curves}

\begin{figure}[!ht]
    \centering
    \makebox[\textwidth][c]{%
    \begin{minipage}{\textwidth}
        \centering
        \subfloat[Validation weighted F1 score]{\includegraphics[scale=0.3]{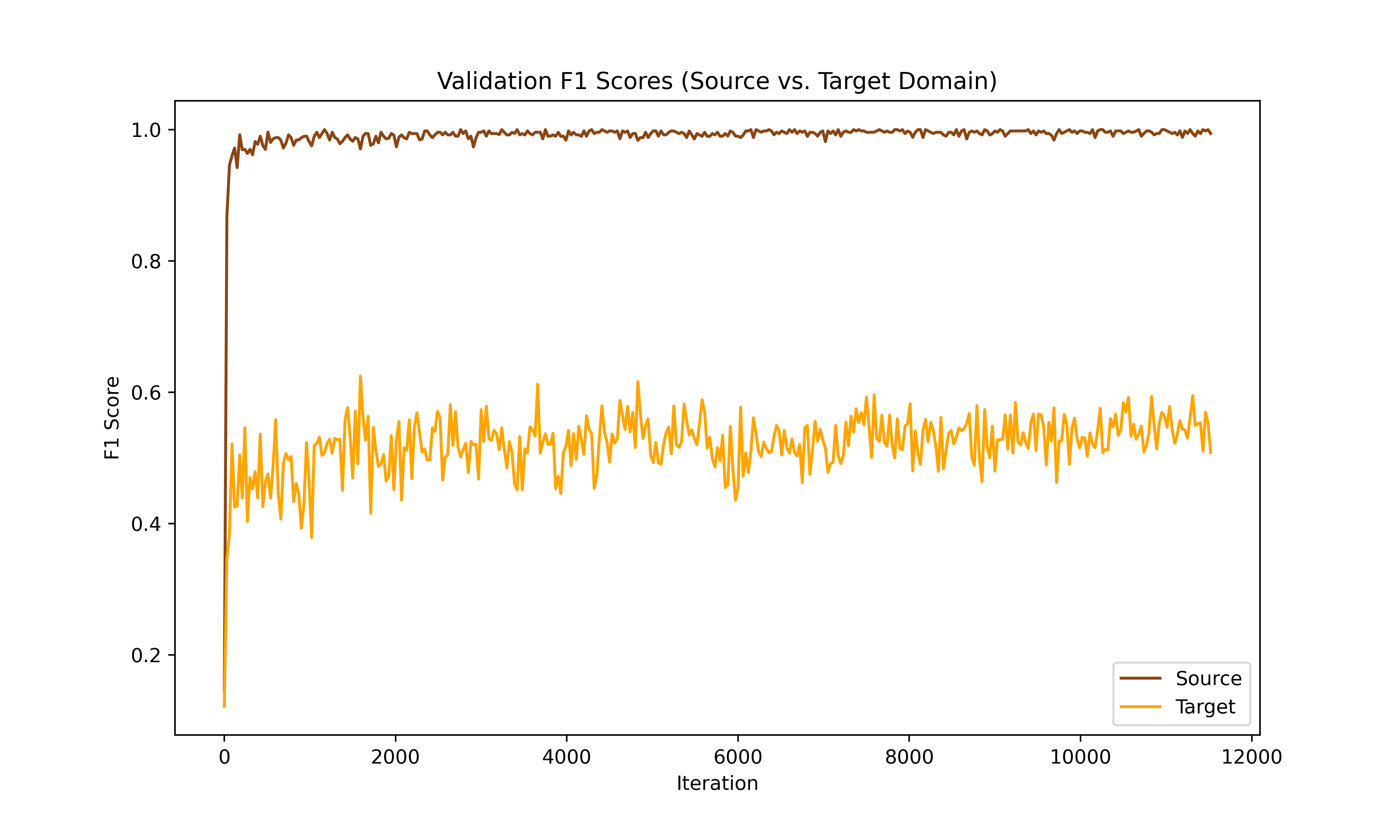}}
        \qquad
        \subfloat[Source Domain Validation Loss]{\includegraphics[scale=0.3]{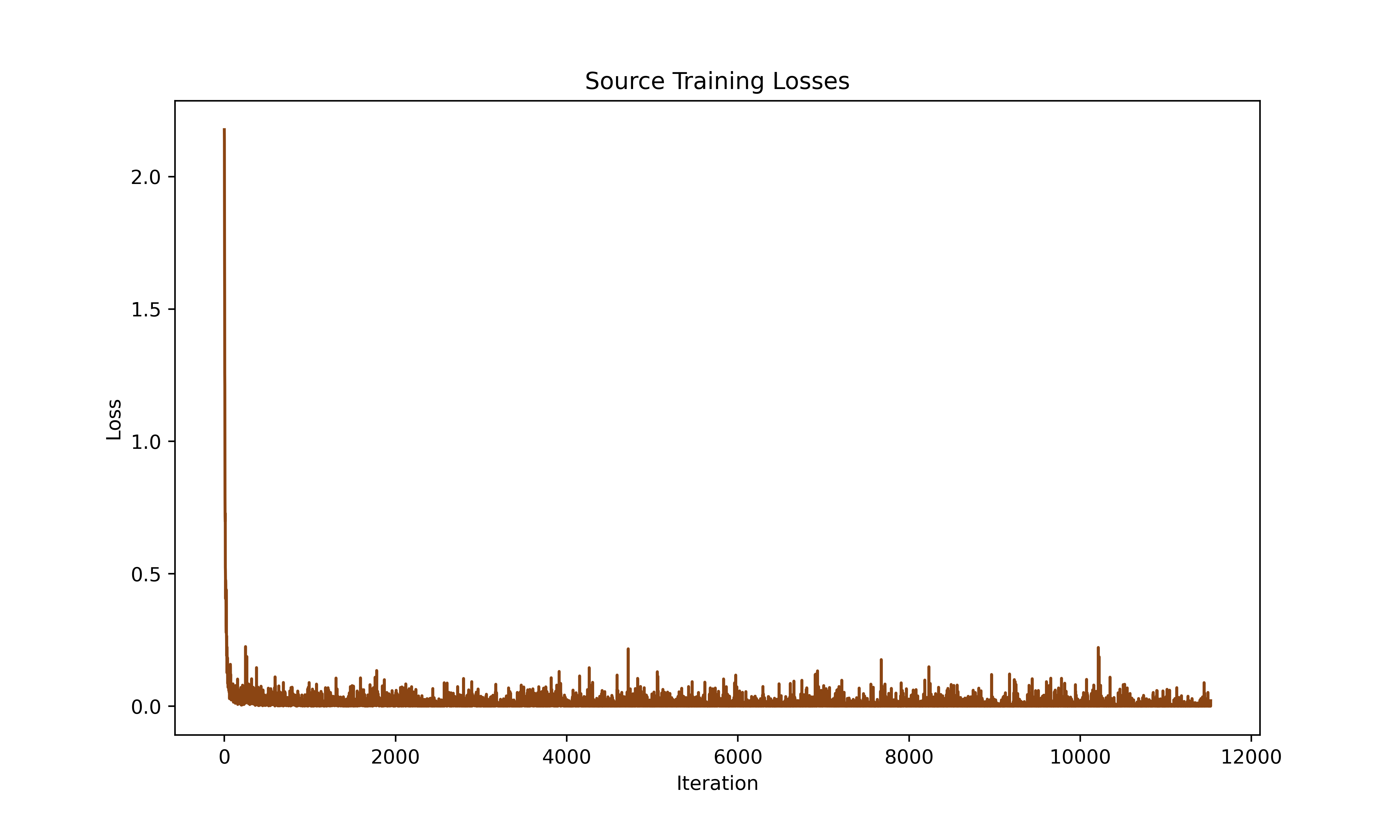}}
        \qquad
        \subfloat[Validation Accuracies]{\includegraphics[scale=0.3]{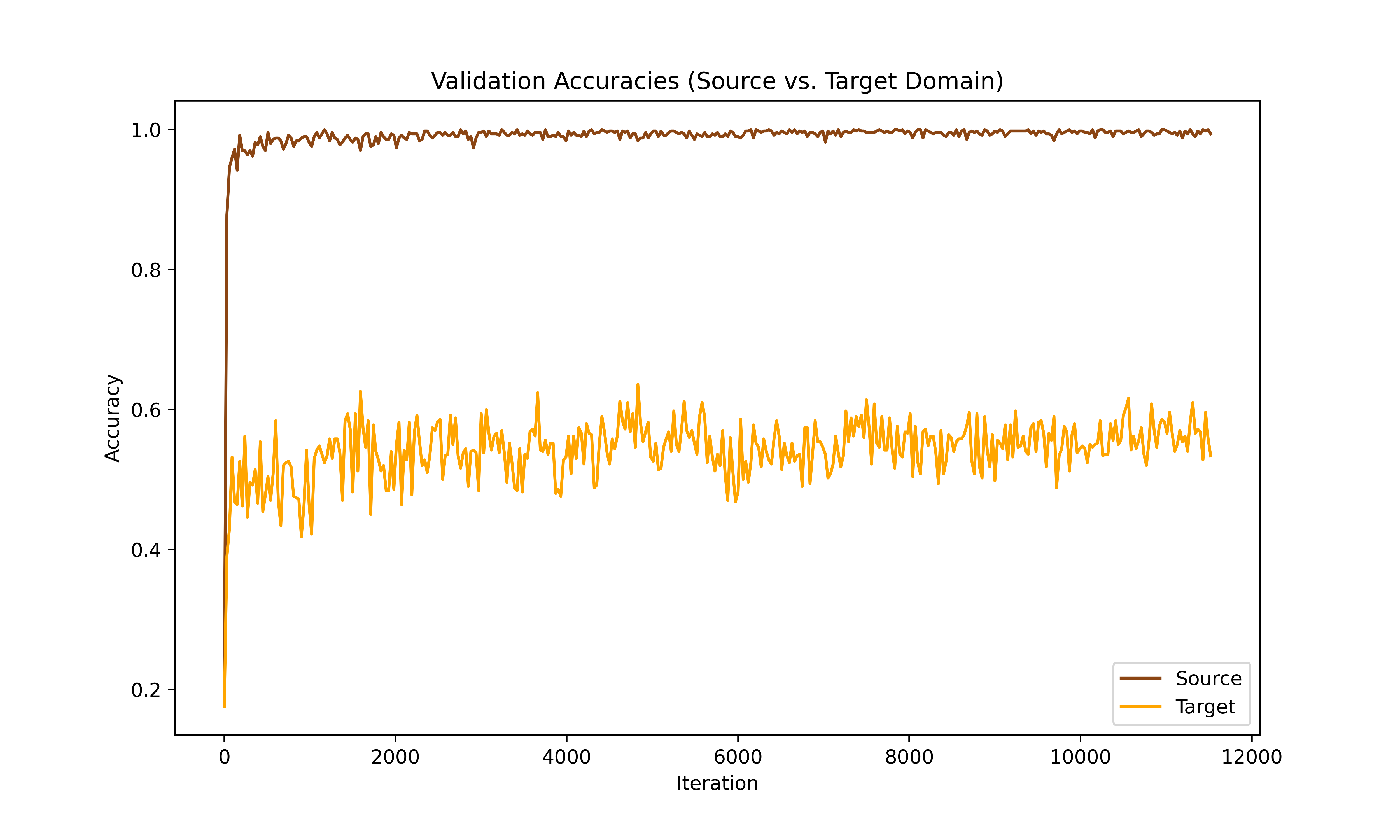}}
    \end{minipage}}
    \caption{Final Base-Source model validation curves.}
    \label{fig:BaseSourceValidation}
\end{figure}

Similarly to the DANN model, the metrics from section \ref{paragraph:Base_validation} were plotted for the combination of hyperparameters that performed best on the validation set. They can be found in the form of a table in Table \ref{tab:BaseSourceHP}. The corresponding plots are depicted in Figure \ref{fig:BaseSourceValidation}.

During the training process, the validation accuracy on the target domain stagnated and oscillated around 0.5, while the validation error on the source domain decreased to nearly 0. This indicates that the model struggled to adapt to the target domain, and proves the need for domain adaptation to solve the problem at hand.

\subsection{CORAL model}

\subsubsection{Hyperparameter tuning results}
\label{CORAL_results}

The results of the fine-tuning of the CORAL model are shown in Figure \ref{tab:CORALHP}. This hyperparameter tuning was done with the following fixed parameters: $Batch\;size = 200$ ; $Number\;of\;epochs = 30$ . The best validation accuracy value on the target dataset was 85.4\%, found with $\gamma \text{ (Focal Loss)} = 5$, $Learning\;rate = 0.0005$, and $\lambda_{max} \text{ (coral)}=1$.

\begin{table}[htbp]
\centering
\begin{tabular}{cccc}
\toprule
\textbf{$\gamma$ (Focal Loss)} & \textbf{Learning Rate} & \textbf{$\lambda_{max}$ (DA factor)} & \textbf{Best Validation Accuracy} \\ \midrule
2 & 0.001 & 0.1 & 80.9\% \\
2 & 0.001 & 1 & 83.2\% \\
2 & 0.001 & 10 & 74.7\% \\
2 & 0.0005 & 0.1 & 80.7\% \\
2 & 0.0005 & 1 & 83.6\% \\
2 & 0.0005 & 10 & 84.2\% \\
5 & 0.001 & 0.1 & 81.1\% \\
5 & 0.001 & 1 & 81.4\% \\
5 & 0.001 & 10 & 83.5\% \\
5 & 0.0005 & 0.1 & 82.5\% \\
\textbf{5} & \textbf{0.0005} & \textbf{1} & \textbf{85.4\%} \\
5 & 0.0005 & 10 & 84.4\% \\
\bottomrule
\end{tabular}
\caption{CORAL hyperparameter tuning results}
\label{tab:CORALHP}
\end{table}

As a reminder, $\lambda_{max}$ is the maximum value of the domain adaptation factor reached through a linear ascent starting at 0, as explained in Section \ref{paragraph:CORAL_validation}.

The drastic difference in validation accuracies between the top and bottom 4 rows of the table can clearly be attributed to the learning rate, showing that 0.01 was overshooting. Furthermore, looking solely at the top 4 rows, we can see that the $\lambda_{max}$ values of 0.01 and 100 were too small and too large respectively, while 10 was the domain adaptation factor value that struck the best balance between minimizing the domain divergence and learning to classify the pieces to a sufficient degree.

\subsubsection{Validation curves}

\begin{figure}[!ht]
    \centering
    \makebox[\textwidth][c]{%
    \begin{minipage}{\textwidth}
        \centering
        \subfloat[Validation weighted F1 score]{\includegraphics[scale=0.3]{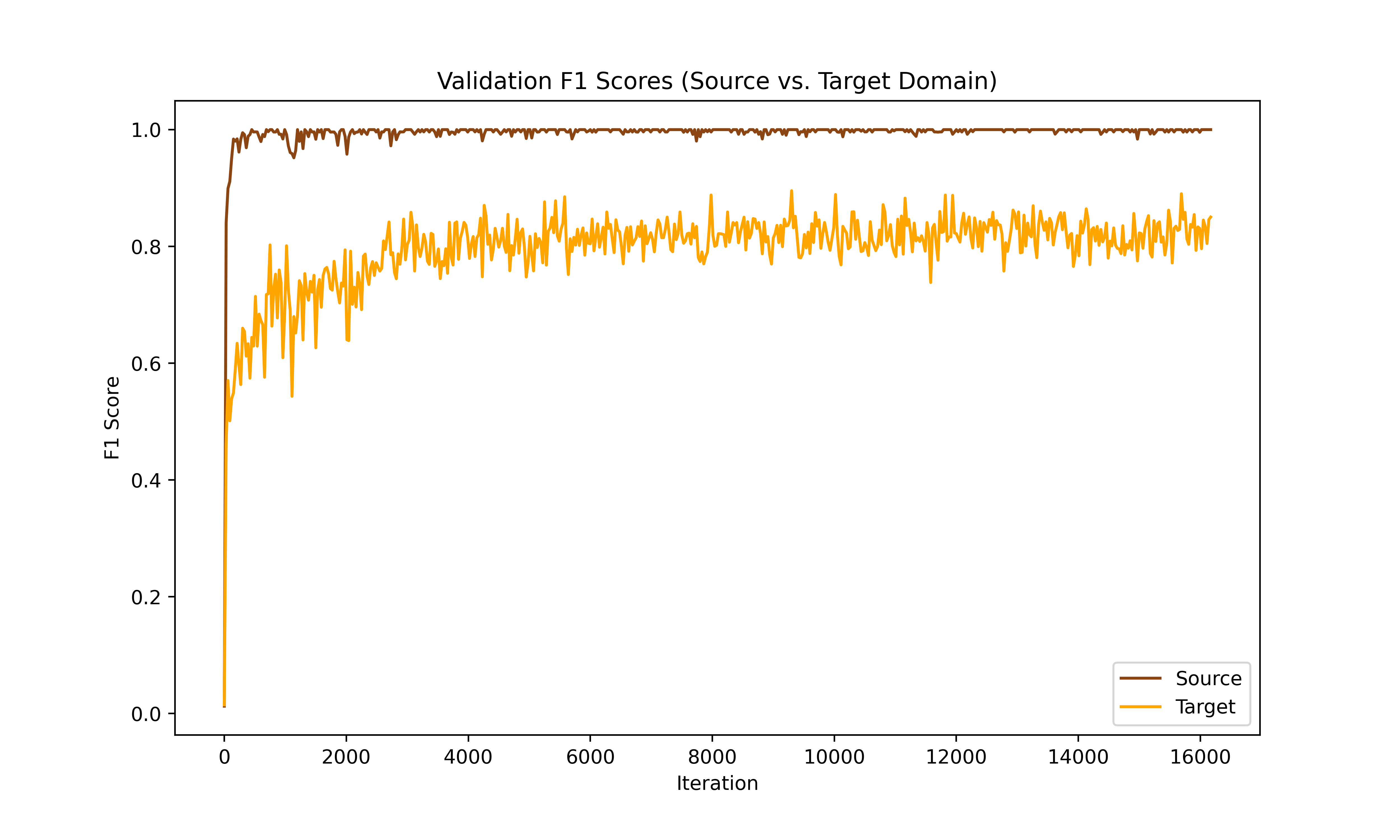}}
        \qquad
        \subfloat[Validation total and CORAL losses]{\includegraphics[scale=0.3]{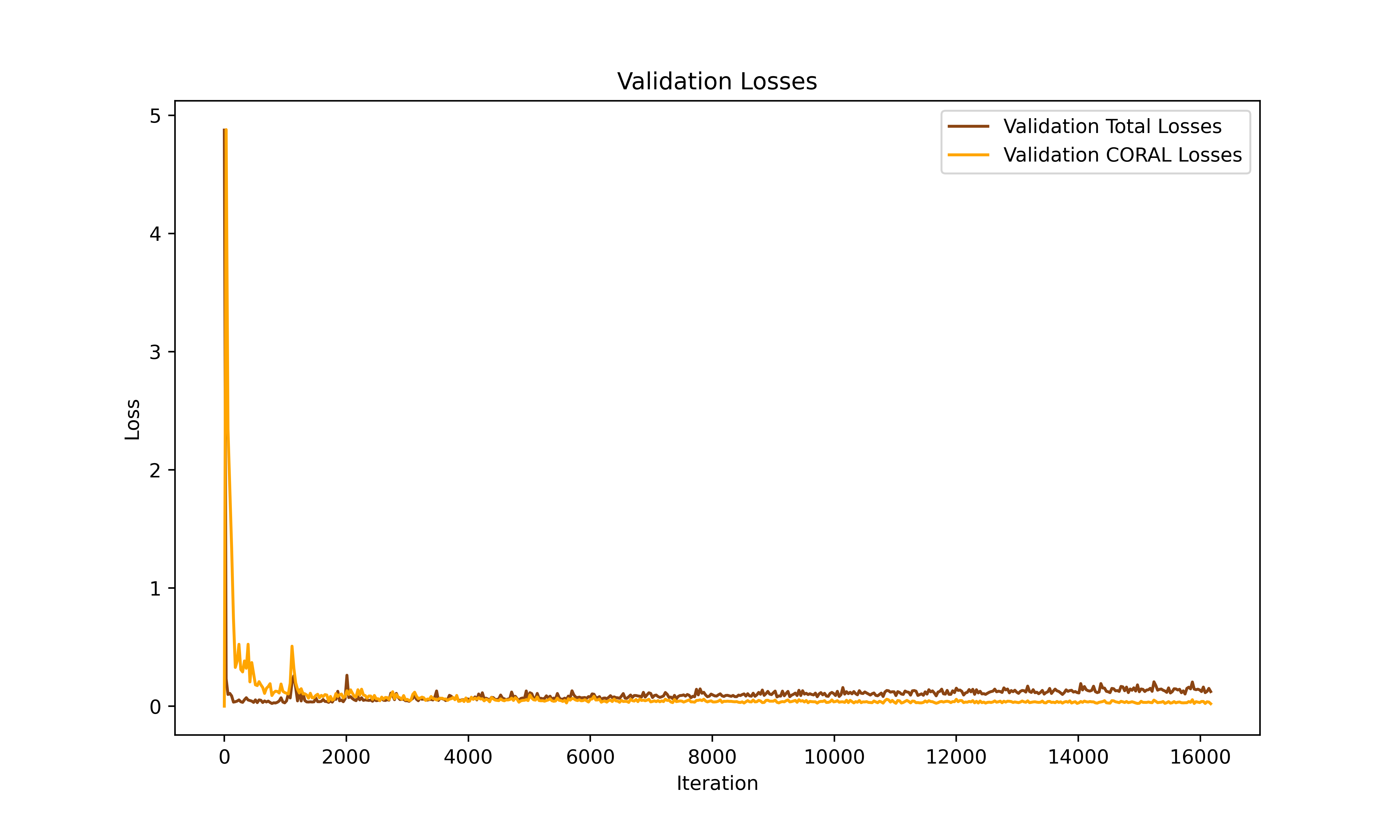}}
        \qquad
        \subfloat[Validation Accuracies]{\includegraphics[scale=0.3]{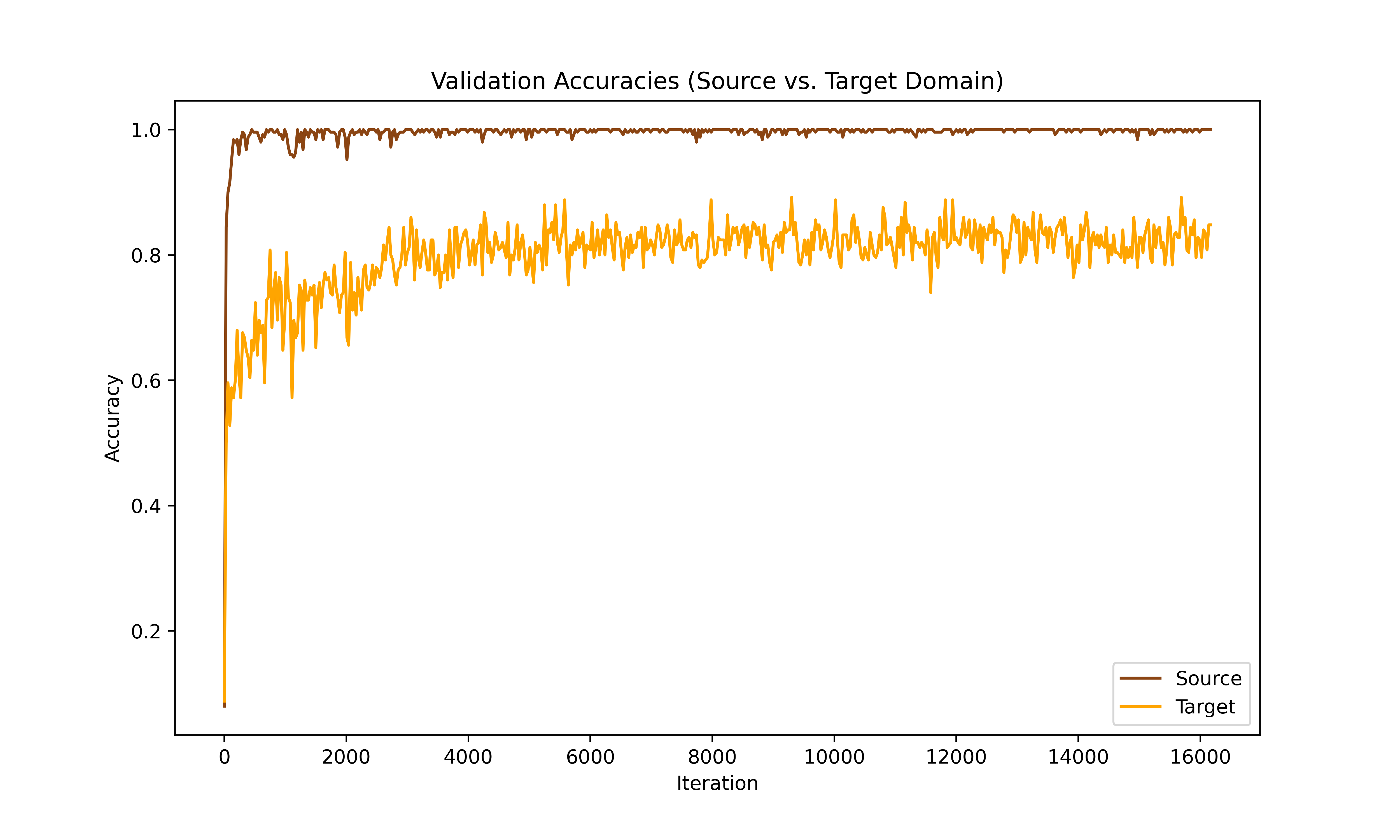}}
        \qquad
        \subfloat[Validation source and target domain classification losses]{\includegraphics[scale=0.3]{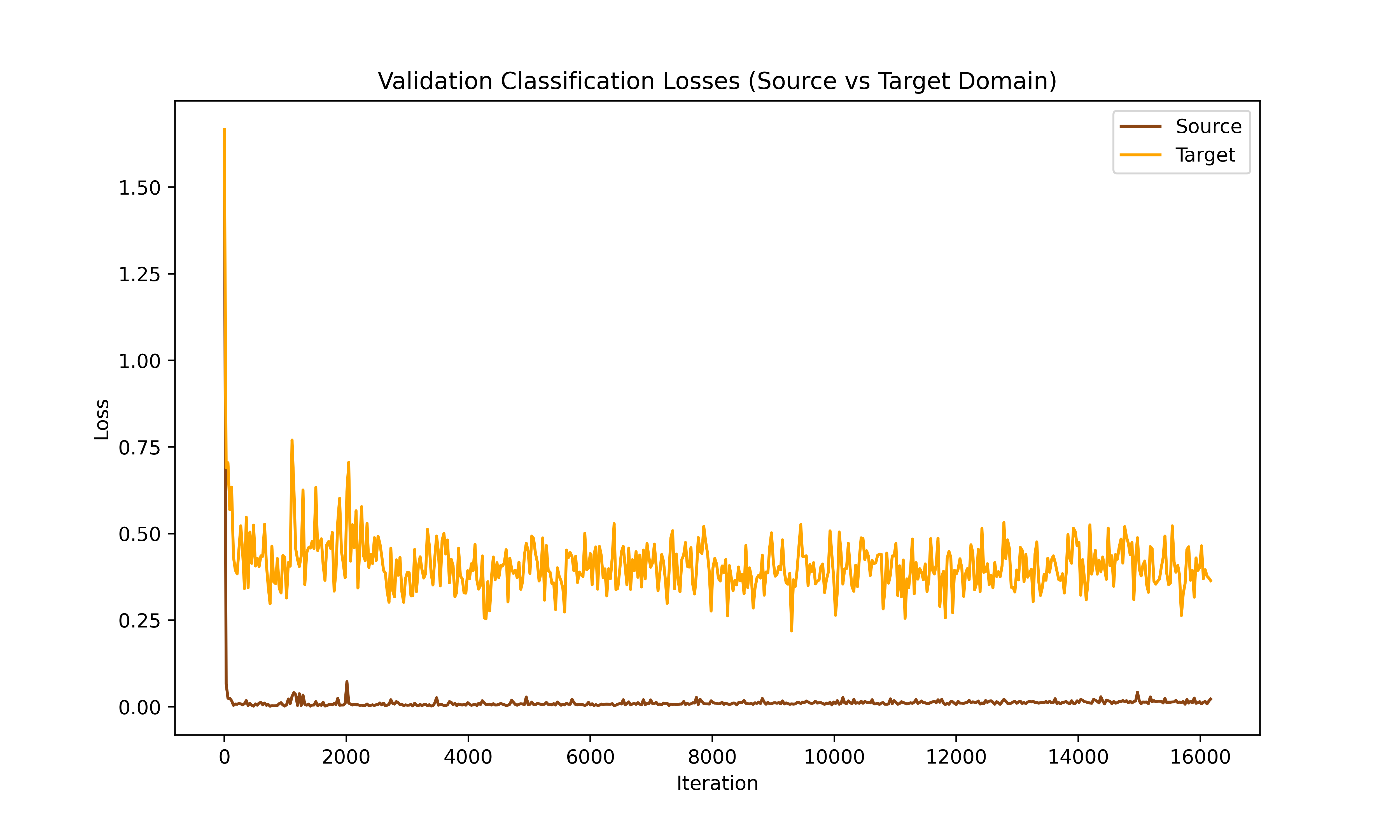}}
    \end{minipage}}

    \caption{Final CORAL Model validation curves.}
    \label{fig:CORALValidation}
\end{figure}

Similarly as the previous models, during the training of the CORAL model, the validation metrics in \ref{paragraph:CORAL_validation} were monitored thoughout iterations. Figure \ref{fig:CORALValidation} shows these metrics for the hyperparameter combination with the best validation accuracy that can be found in Table \ref{tab:CORALHP}.

As the regularization parameter $\lambda$ increased throughout iterations from 0 to $\lambda_{max}$, we observe that the CORAL loss showed a significant decrease in value (see Figure \ref{fig:CORALValidation}b). This can be attributed to the fact that as $\lambda$ increases, the weight of the CORAL loss in the total loss becomes more prominent, leading to a more effective alignment of the domain features. Because a larger $\lambda$ places more emphasis on aligning the feature distributions across domains, which may result in a higher overall loss value, the total loss increased with the number of iterations as seen in \ref{fig:CORALValidation}b. However, this increase in loss is not an indicator of overfitting of any kind, and rather a natural result of increasing the value of $\lambda$ throughout iterations, which modifies the scale of the total loss. Further, the classification loss for the source domain increased throughout iterations, while the target domain classification loss remained roughly constant (see Figure \ref{fig:CORALValidation}d). This can be attributed to the fact that, as $\lambda$ becomes larger, the classification loss is assigned a smaller portion of the total loss, causing the backpropagation process to focus less on optimizing classification accuracy and more on optimizing the CORAL loss. However, it is important to note that, even though the source domain classification loss increased, this is totally normal for an incremental domain adaptation approach, since as epochs progress, the aim becomes less and less to classify the source data, and the focus shifts to the target data. This is justified by the fact that the target domain accuracies and F1 scores consistently improved (See Figure \ref{fig:CORALValidation}a and \ref{fig:CORALValidation}c). This indicates that the domain alignment achieved through the CORAL loss was very beneficial for improving the model's performance on the target data, despite the increases in the source domain losses.

\subsection{Base-Target model}

\subsubsection{Hyperparameter tuning results}
\label{subsubsection:Base_results_target}

The results of the fine-tuning of the Base model trained on the target domain data directly are shown in Figure \ref{tab:BaseTargetHP}. This hyperparameter tuning process was executed using the following fixed parameters: $Learning\;rate = 0.001$, $Batch\;Size=100$, and $Number\;of\;epochs = 5$. The best validation accuracy value on the target dataset was 94.9\%, found with $\gamma= 5$,  and $Dropout\;rate=0.2$. Again, we remind the reader that this model is "cheating" by looking at the labels of the target domain data, and is only included for comparison purposes.

\begin{table}[htbp]
\centering
\begin{tabular}{ccc}
\toprule
\textbf{$\gamma$ (Focal loss)} & \textbf{Dropout Rate} & \textbf{Best Validation Accuracy} \\ \midrule
2 & 0 & 94.7\% \\
2 & 0.2 & 93.3\% \\
2 & 0.5 & 93.8\% \\
5 & 0 & 94.7\% \\
\textbf{5} & \textbf{0.2} & \textbf{94.9\%} \\
5 & 0.5 & 93.8\% \\
\bottomrule
\end{tabular}
\caption{Base-target hyperparameter tuning results}
\label{tab:BaseTargetHP}
\end{table}

\subsubsection{Validation curves}

\begin{figure}[!ht]
    \centering
    \makebox[\textwidth][c]{%
    \begin{minipage}{\textwidth}
        \centering
        \subfloat[Validation weighted F1 score]{\includegraphics[scale=0.3]{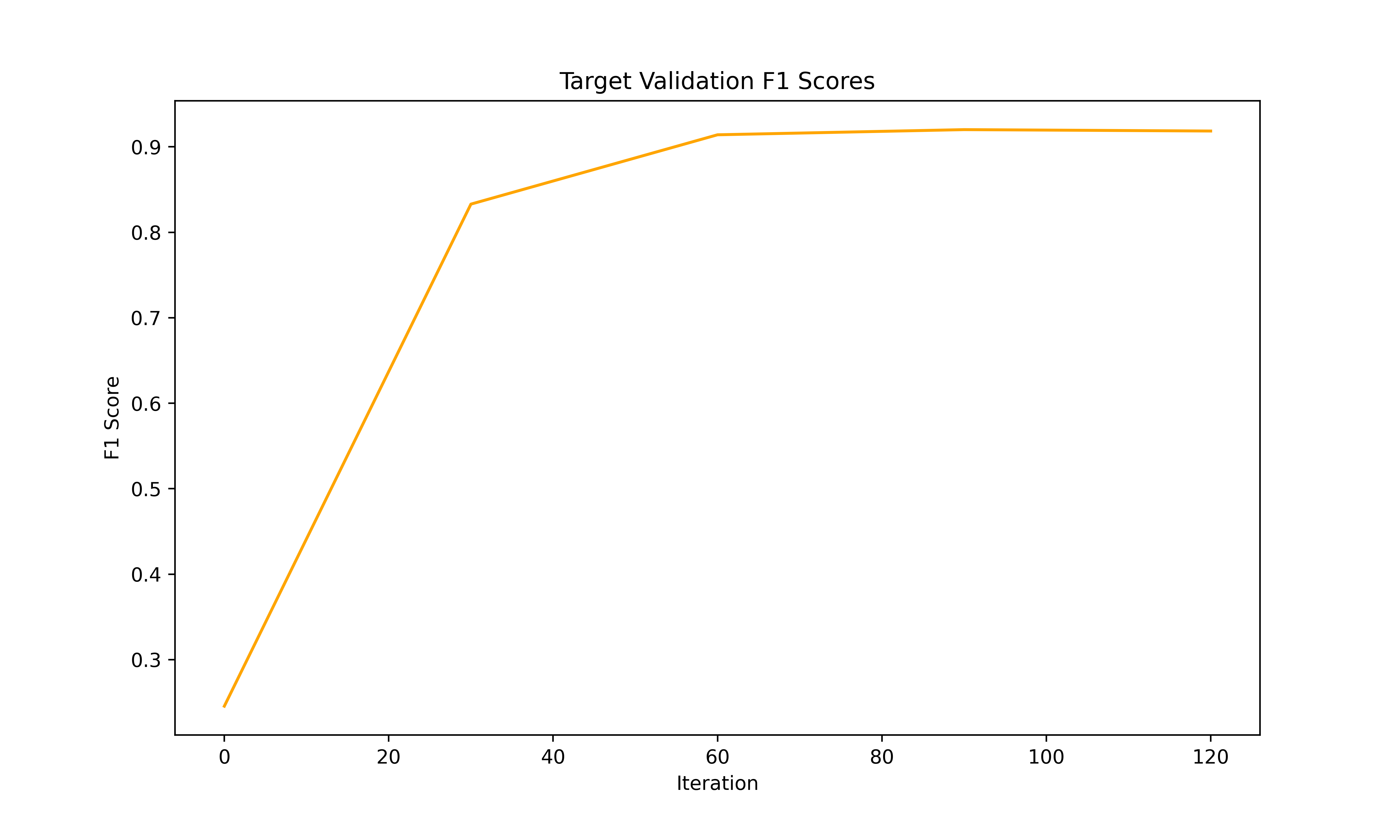}}
        \qquad
        \subfloat[Validation Losses]{\includegraphics[scale=0.3]{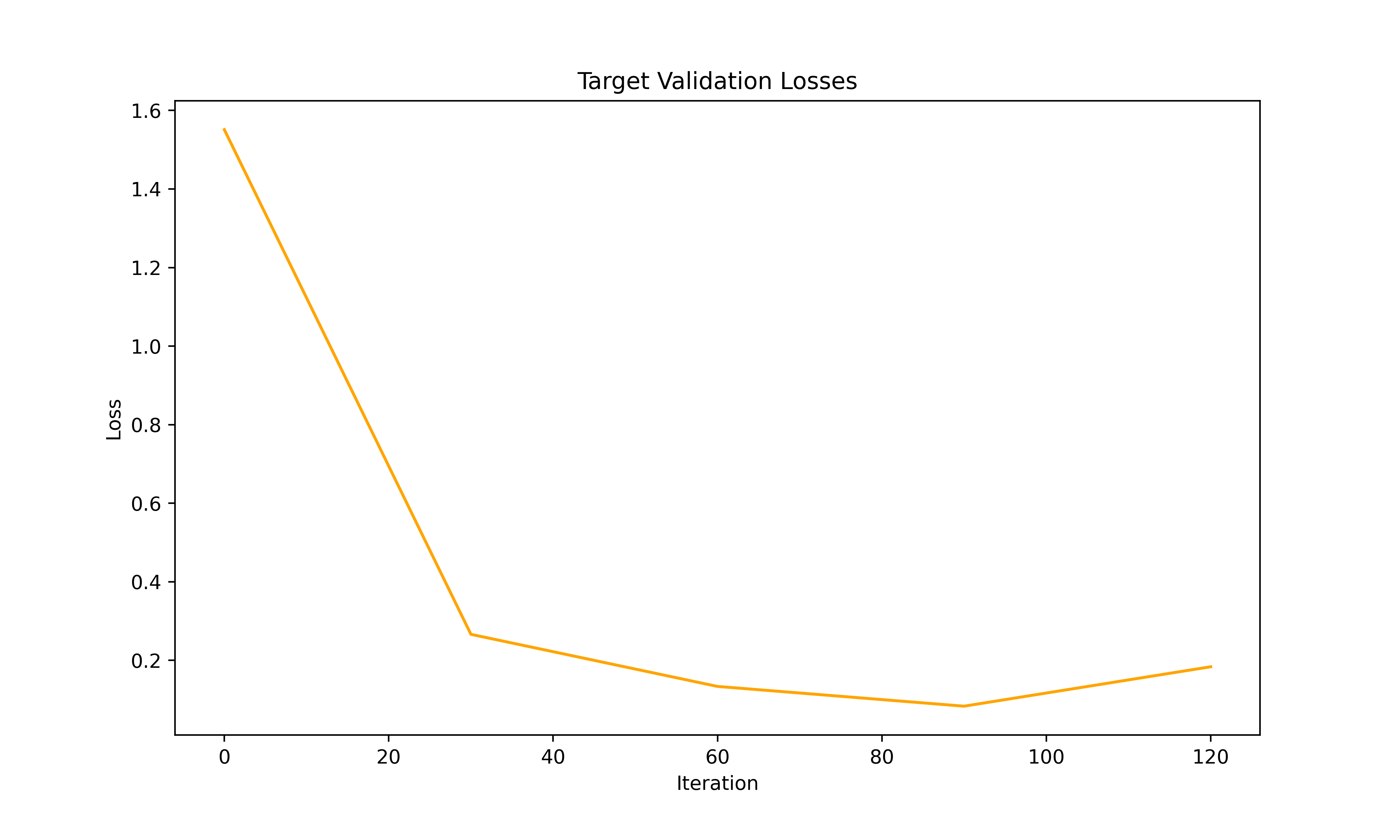}}
        \qquad
        \subfloat[Validation Accuracies]{\includegraphics[scale=0.3]{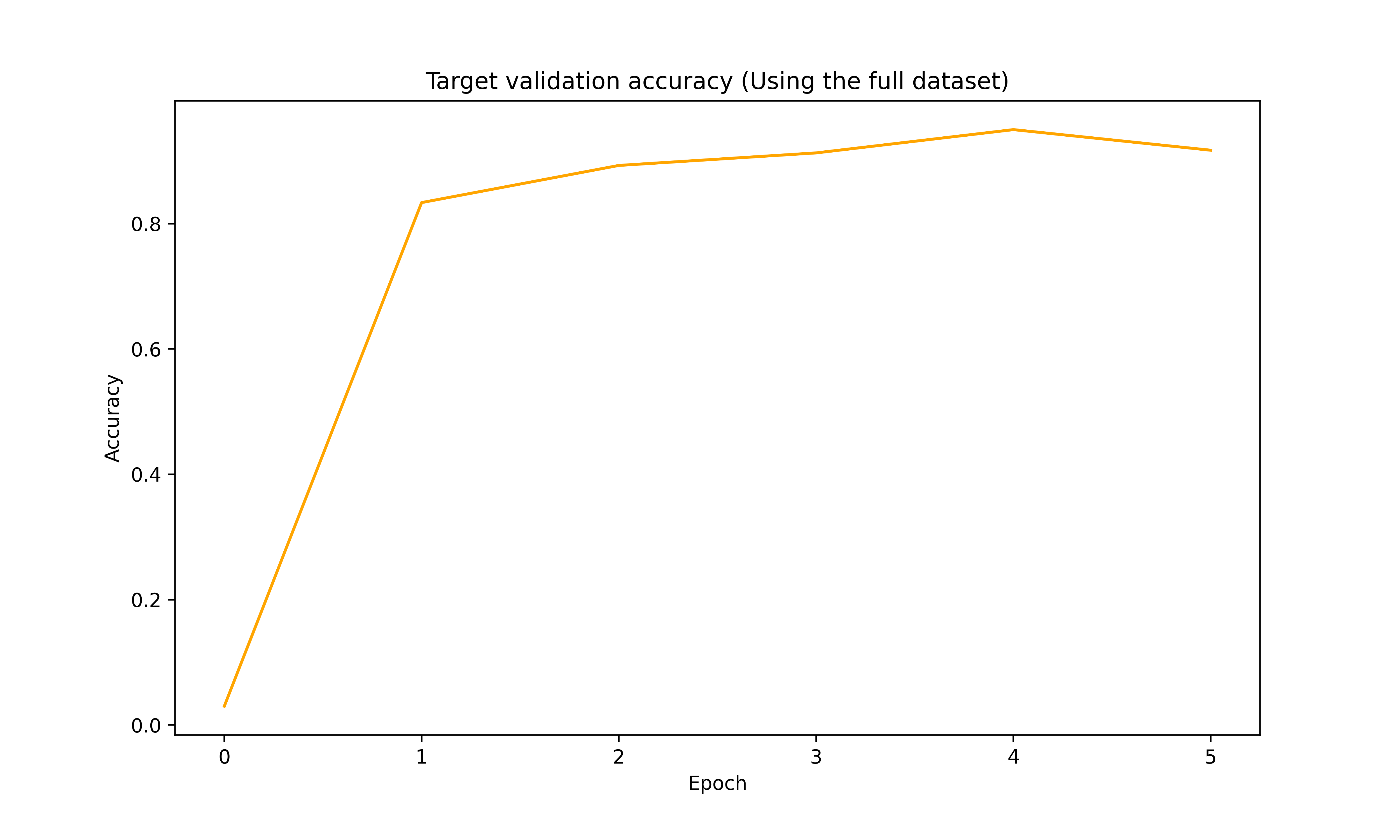}}
    \end{minipage}}
    \caption{Final Base-Target Model validation curves.}
    \label{fig:BaseValidation}
\end{figure}

Similarly to the base model trained on the source data, the metrics from section \ref{paragraph:Base_validation} were plotted for the combination of hyperparameters that performed best on the validation set. They can be found in Table \ref{tab:BaseTargetHP}. The corresponding plots are depicted in Figure \ref{fig:BaseValidation}.

We can see that the model starts to overfit towards the end of the training, which is why our code checkpointed every validation iteration and reminded to the one that yielded the highest accuracy. Otherwise, this benchmark model almost perfectly learns to differentiate the domains

\subsection{Cross-model testing metric comparison}

For this section, hyperparameter combinations that yielded the best validation accuracies were utilized for each of the models. Then, multiple metrics were accumulated using the labeled target and source domain testing sets. However, it is to be noted that all reported metrics were achieved on the \textbf{oversampled and augmented} testing sets such that all classes are balanced. However, in a realistic chess classification setting, the proportion of empty squares would be much larger than any other class, which means that the accuracy that the proposed model would achieve in a real-life situation would be much larger. In any case, only the balanced dataset accuracies are reported because they provide more insight into the model's performance.

\subsubsection{Testing metrics}

The results obtained after testing the three implemented models showed varying levels of accuracy across the two domains. The metrics measured are the standard testing accuracy, the F1-score, and the average AUPRC, which measures the overall quality of the model's predictions, taking into account both precision and recall across all possible classification thresholds \cite{PRCPaper}.

On the source domain data, the CORAL model achieved the highest testing accuracy of 99.94\%. Table \ref{tab:TestACCBalancedSource} shows the balanced source data final testing metrics. This table is interesting, even though it does not really represent the objective of this paper.

\begin{table}[htbp]
\centering
\begin{tabular}{lccc}
\toprule
\textbf{Model} & \textbf{Testing Accuracy} & \textbf{F1-Score} & \textbf{Average AUPRC} \\ \midrule
Base-source & 99.17\% & 0.992 & 0.992 \\
\textbf{CORAL} & \textbf{99.94\%} & \textbf{0.999} & \textbf{0.999} \\
DANN & 99.68\% & 0.997 & 0.997 \\
Base-target & 36.56\% & 0.340 & 0.429 \\
\bottomrule
\end{tabular}
\caption{Testing metrics on the balanced source domain.}
\label{tab:TestACCBalancedSource}
\end{table}

More importantly, when it comes to the balanced target domain data, the CORAL model achieved the highest accuracy of 85.43\%, of course disregarding the base model trained on the target data since it can see the target domain labels. The superior performance of the DANN and CORAL models compared to the base model trained on the source data highlights the effectiveness of domain adaptation as a viable solution for addressing domain shift and improving model accuracy in real-life data scenarios. Table \ref{tab:TestAccTargetBalanced} displays the testing metrics on the target domain data.

\begin{table}[htbp]
\centering
\begin{tabular}{lccc}
\toprule
\textbf{Model} & \textbf{Testing Accuracy} & \textbf{F1-Score} & \textbf{Average AUPRC} \\ \midrule
Base-source & 57.59\% & 0.553 & 0.595 \\
\textbf{CORAL} & \textbf{85.43\%} & \textbf{0.852} & \textbf{0.859} \\
DANN & 83.59\% & 0.832 & 0.842 \\
Base-target & 93.00\% & 0.930 & 0.937 \\
\bottomrule
\end{tabular}
\caption{Testing metrics on the balanced target domain data.}
\label{tab:TestAccTargetBalanced}
\end{table}

Finally, the most important metric in this project is the testing accuracy on the imbalanced target domain data, since it reflects the percentage of squares that will accurately be classified in a real-life setting where empty squares dominate the distribution. Table \ref{tab:TestAccTargetImbalanced} shows that the DANN model outperforms the other models, again disregarding the base model trained on target data since it is only included for comparison purposes.

\begin{table}[hbp]
\centering
\begin{tabular}{lccc}
\toprule
\textbf{Model} & \textbf{Testing Accuracy} & \textbf{F1-Score} & \textbf{Average AUPRC} \\ \midrule
Base-source & 87.43\% & 0.874 & 0.556 \\
CORAL & 89.91\% & 0.919 & 0.825 \\
\textbf{DANN} & \textbf{92.31\%} & \textbf{0.932} & \textbf{0.793} \\
Base-target & 95.18\% & 0.956 & 0.889 \\
\bottomrule
\end{tabular}
\caption{Testing metrics on the imbalanced target domain data.}
\label{tab:TestAccTargetImbalanced}
\end{table}

It is interesting to note that, while the DANN model outperforms the CORAL model based on the accuracy and F1-score metrics, the average AUPRC of the CORAL model turned out to be higher than that of the DANN model.

Furthermore, we can clearly see that the DANN model performs very similarly to the Base-target model which is an upper bound on its performance, showing that our DA process was pushed to its limit. Indeed, the percentage of accuracy lost through domain adaptation is:

\[
    \frac{Acc_{Base-Target} - Acc_{DANN}}{Acc_{Base-Target}} = \frac{95.18 - 92.31}{95.18} = 3.02 \%
\]

It is then a question of whether this 3\% loss in accuracy is worth the labeling effort saved by this technique.

\subsubsection{Confusion matrices}

Four confusion matrices were generated based on the balanced target domain data, one for each model. We only performed this process for the balanced target domain data since we are comparing each class separately. It is worth noting that these matrices are normalized by row, so going through each, row by row, gives an idea of what the model classified the row piece as. A perfect model should have a confusion matrix with a value of 100\% in every diagonal entry, meaning that each class is predicted as itself. 

\begin{figure}[!ht]
    \centering
    \makebox[\textwidth][c]{%
    \begin{minipage}{1\textwidth}
        \centering
        \subfloat[Base-Source Model]{\includegraphics[scale=0.3]{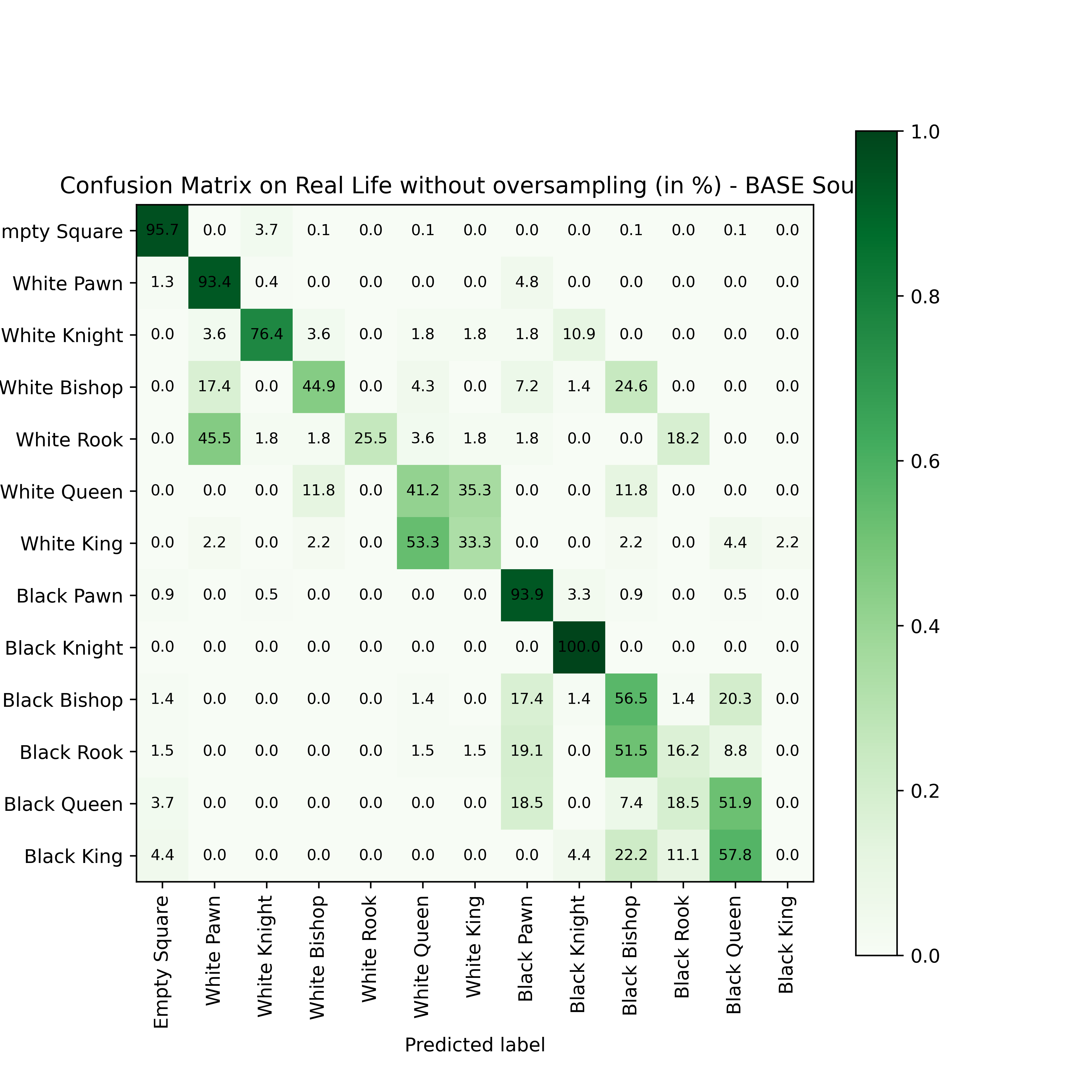}}
        \qquad
        \subfloat[DANN]{\includegraphics[scale=0.3]{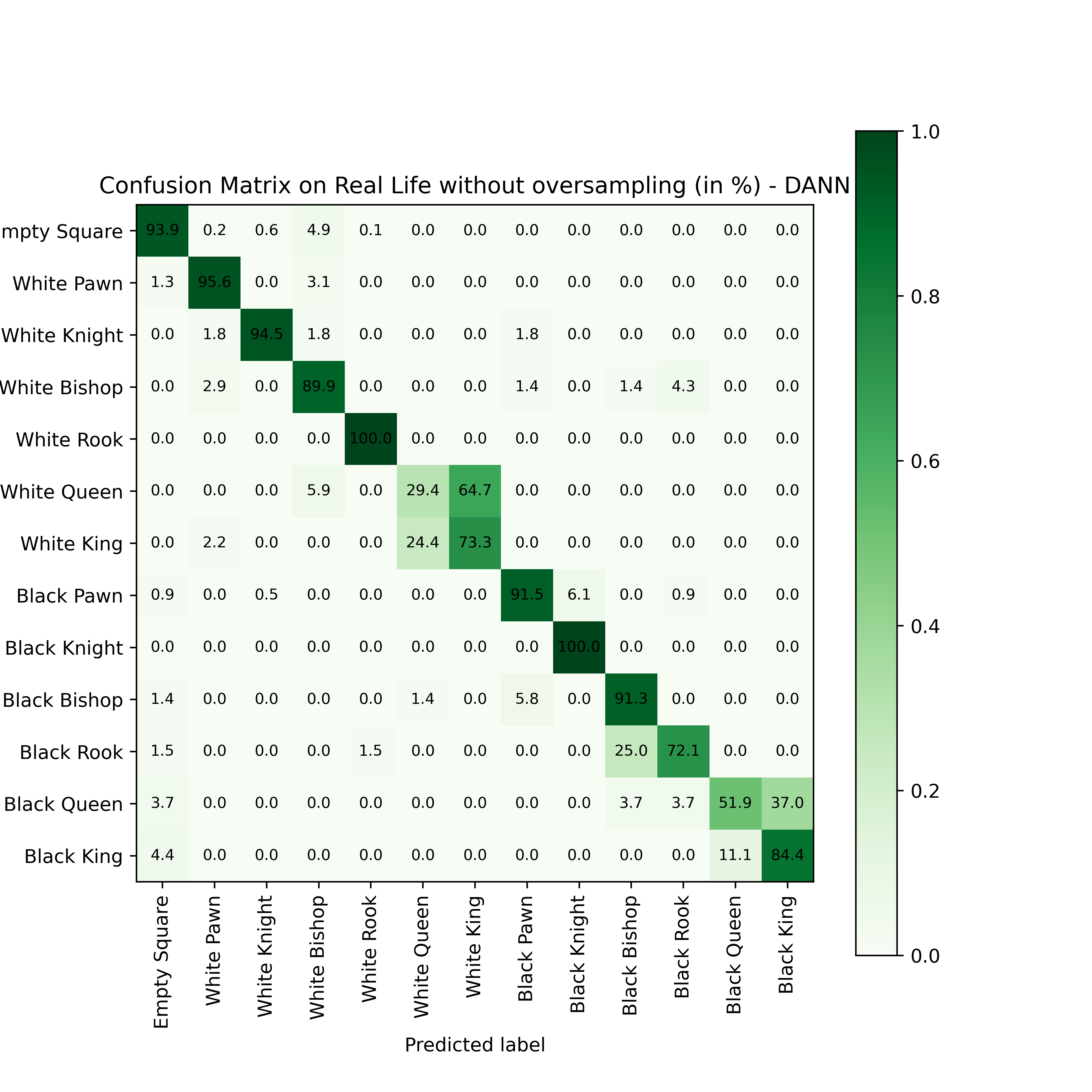}}
        \qquad
        \subfloat[CORAL]{\includegraphics[scale=0.3]{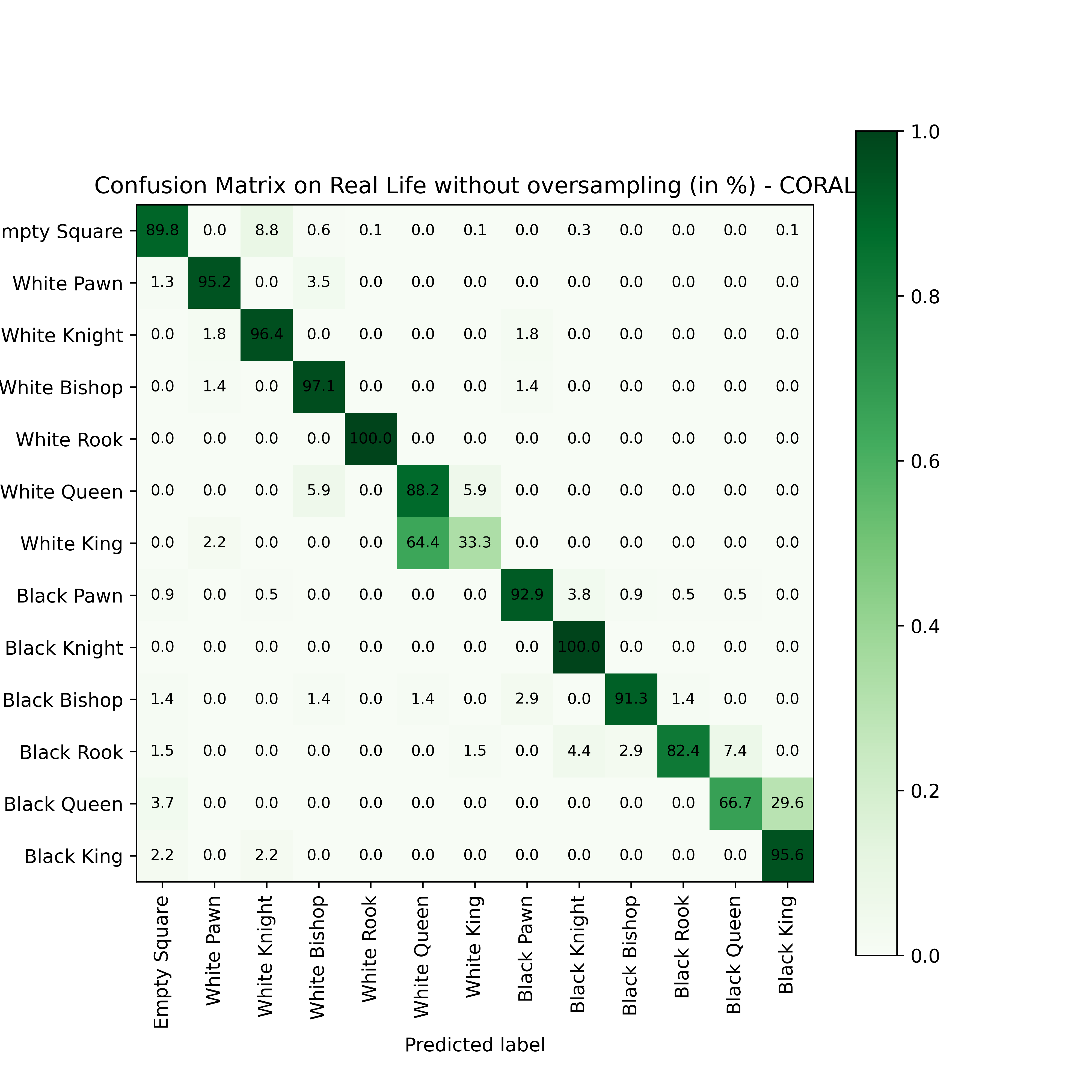}}
        \qquad
        \subfloat[Base-Target Model]
        {\includegraphics[scale=0.3]{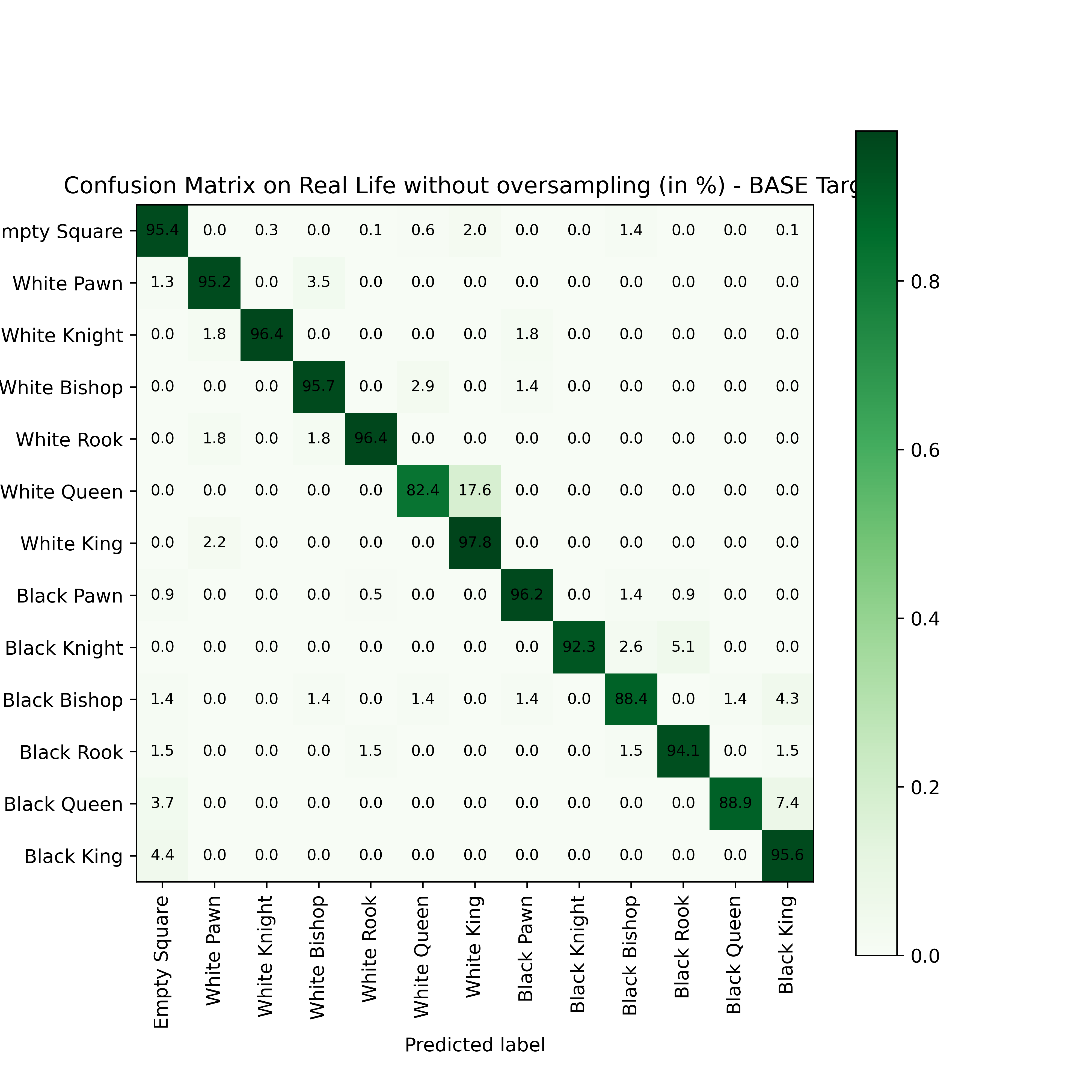}}
    \end{minipage}}
    \caption{Confusion matrices on the balanced target dataset}
    \label{fig:MATRICES}
\end{figure}

The Base-Source model's confusion matrix shown in Figure \ref{fig:MATRICES}a is not very diagonal, indicating that it misclassifies certain pieces. For example, it exhibits notable confusion between chess pieces that have similar top-view appearances. Indeed, it confuses a significant percentage of bishops with pawns. Additionally, it consistently confuses kings and queens.

On the other hand, both DANN and CORAL models show much more diagonal confusion matrices, highlighting once again that domain adaptation was the solution to this unsupervised learning problem. The DANN confusion matrix is depicted in Figure \ref{fig:MATRICES}b, while the CORAL confusion matrix is presented in Figure \ref{fig:MATRICES}c.

Interestingly, both models showed some degree of confusion between kings and queens, with the CORAL baseline showing slightly more confusion than the DANN model. This could explain why DANN achieved higher accuracy results compared to CORAL, as it may be making relatively fewer mistakes in classifying kings and queens accurately. In any case, it is clear that these models would perform exceptionally well on a more realistic chess data distribution, since all chess games have 1 king and usually at most 1 queen (except for rare cases where piece promotions are achieved).

Finally, the Base-Target model performs exceptionally well as expected, with most of its error inherent to predicting white queens as white kings.

\section{Conclusions and Discussion}

\subsection{Summary}

In conclusion, this paper presents an end-to-end pipeline that utilizes domain adaptation to automate the process of predicting the labels of chess pieces in unlabeled top-view images of chess boards. The pipeline comprises a pre-processing phase, a Deep Learning model for prediction, and a post-processing phase for generating FEN strings and reconstructing the image. Synthetic 3D images of chessboards were used as the source domain data, while unlabeled top-view photographs of chess positions served as the target domain data. Three approaches for optimizing the model were considered, including a pre-trained VGG16 Base-Source model trained directly on the source domain data, an improved CORAL model including a CORAL loss for domain adaptation, and a Domain Adversarial Neural Network (DANN) with adversarial training. The Source-Target model, which was trained directly on the target domain data with its label, is also included for comparison purposes.

The testing metrics were evaluated for the various models using both balanced and imbalanced target domain data. The results showed that the DANN model was the most accurate among the models when looking at the imbalanced target domain data which most reflects the real scenario performance, with the CORAL model being a close contender. In contrast, the Base-Source model performed poorly in comparison to both. These findings highlight the potential of domain adaptation, for automating the time-consuming process of manual record-keeping in chess without requiring any data labeling.

\subsection{Limitations}

\subsubsection{Limitation to top-view images}
The end-to-end pipeline, as discussed in Section \ref{subsubsection:Target_preprocessing}, is specifically designed to classify top-view images of the target dataset. However, this approach has certain limitations. Firstly, the pipeline partially tolerates variations in angles. While it can successfully rectify images that are not entirely top-view through image warping, it may struggle with accurately rectifying side-view images. Additionally, the model may encounter challenges in distinguishing certain chess pieces, such as bishops and pawns, which have similar appearances when viewed from the top. To address this issue, future research could involve modifying the warping and square detection algorithms in order to support training on side-view images, which would make the model completely robust to camera angles, as mentioned in Section \ref{paragraph:angle_invariance}.

\subsubsection{Dependence on board texture and piece set}

During training, validation, and testing, the models were trained using the same target domain chess set and its corresponding source domain imitation. This means that they inherently lack the ability to perform consistently on different board textures and piece styles. To improve the models for recognition across a range of chess sets in future work, it would be beneficial to attempt an unsupervised domain adaptation task where the source and target domain datasets mix a wide range of board textures and piece styles.

\subsection{Advantages}

\subsubsection{Excellent results through the use of domain adaptation}

The results compared in Section \ref{sec:Results} clearly show the superiority of the DA approaches when compared to the Base-Source baseline which utilized no DA. Further, the confusion matrices from said section showed that most pieces were classified with near-perfect accuracy, except for the kings and queens which make up the minority of a real-life distribution. This showcases that it is worth considering an unsupervised domain adaptation approach over labeling the target data and training the model directly on it. Through such an approach, this project was able to meet all of the constraints mentioned in the introduction, from having the generated data be similar enough to the target domain data, to surpassing the performance of the Base model, thus allowing us to consider the project as successful.

\subsubsection{Abundance of training examples without manual labeling}

This project highlights that generating data and performing domain adaptation can yield comparable results to labeling and training on the target domain, without the effort required for the labeling. Further, the ease of generating data meant that the total size of the training dataset was much larger than what would have been achieved through manual labeling. On the data generation front, future work could be done to have the source domain match the target domain even more closely, which would likely imply even better results. 

\subsubsection{Invariance to rotation, translation, and lighting}

As discussed in Section \ref{paragraph:blender_rendering}, due to the authors having full control over the source domain data, different lighting conditions, top-view camera angles, translations, and rotations were applied to the generated examples. This shows another advantage of data generation, which is the ability to make the trained models invariant to different conditions as required.

\subsection{Future work}

\subsubsection{Other domain adaptation approaches}

A reconstruction-based approach to domain adaptation, such as a Deep Reconstruction-Classification Network (DRCN) \cite{DRCNPaper} could be worth exploring. Such an approach would entail reconstructing the target domain data as the means to achieve domain adaptation.

Another divergence-based approach that could be attempted is the Wasserstein Distance Guided Representation Learning Model (WDGRL) \cite{shen2018wasserstein}, which is a middle ground between the CORAL and DANN models presented in this paper. Essentially, it replaces the domain discriminator portion of the DANN with a domain critic that uses learnable weights to learn the Wasserstein distance between domains. Results presented in the cited paper show a clear improvement over DANN, suggesting that this method is worth investigating.

\subsubsection{Further hyperparameter exploration}

Due to limited time and resources, hyperparameter tuning was not extensively explored in this project. It is possible that the models could have achieved even better performance, had a wider range of hyperparameter combinations been validated. For example, the DANN model was only able to confuse the discriminator to an accuracy level of 0.6 (See Section \ref{subsubsection:DANN_Val_Curves}), something that could have been improved by specifically exploring slightly larger $\lambda$ values.

\subsubsection{Regularization for the domain adaptation models}

Regularization techniques such as weight decay, dropout, and early stopping could be further explored for the two domain adaptation models.

\subsubsection{Live video annotation}

Another potential extension of this research could be processing entire videos to identify various chess positions within them, subsequently utilizing the presented model to analyze each position sequentially. This approach would even further advance the goal of achieving streamlined and readily available automatic chess labeling.

\subsubsection{Utilizing the rules of chess as constraints}

One final thing that could improve the performance of the utilized model is to incorporate the rules of chess as constraints \cite{constraintPaper}. For example, the model could be forced to predict exactly a single king per position during testing, or prohibited from predicting a pawn on the first and eighth ranks, as that is not possible in a real chess position. This could be implemented by adding a soft constraint as a penalty term to the loss function, or even a hard constraint \cite{HardConstraintPaper} incorporated within the network's architecture. In any case, as mentioned in Section \ref{paragraph:blender_rendering}, the generated data, when viewed position by position, fully adheres to every rule of chess. Therefore, anyone willing to attempt such an approach would be able to re-use this project's generated full position dataset before it was cropped into individual squares.

\newpage

\bibliographystyle{IEEEtran}
\bibliography{bibliography}

%%%%%%%%%%%%%%%%%%%%%%%%%%%%%%%%%%%%%%%%%%%%%%%%%%%%%%%%%%%%

\end{document}